%% file: main.tex
\pdfoutput=1

\documentclass[11pt]{article}

\usepackage[final]{acl}

\usepackage{times}
\usepackage{latexsym}

\usepackage[T1]{fontenc}

\usepackage[utf8]{inputenc}

\usepackage{microtype}

\usepackage{inconsolata}

\usepackage{graphicx}

\PassOptionsToPackage{table}{xcolor}
\usepackage{booktabs}   
\usepackage{multirow}
\usepackage{array}      
\usepackage{colortbl}   
\usepackage{caption}    
\usepackage{adjustbox}  
\usepackage{multicol}    
\usepackage{changepage}
\usepackage{listings} 
\usepackage{amsmath}

\newcommand{\dataname}[1]{ChatBench{#1}}
\newcommand{\numQuestions}[1]{396{#1}}
\newcommand{\numAnswers}[1]{144K{#1}}
\newcommand{\numConversations}[1]{7,336}

\newcommand{\gitrepo}[1]{\url{https://github.com/serinachang5/interactive-eval}}
\newcommand{\huggingface}[1]{\url{https://huggingface.co/datasets/microsoft/ChatBench}}

\title{\dataname{}: From Static Benchmarks to Human-AI Evaluation}

\author{Serina Chang \\
  Microsoft Research \\
  University of California, Berkeley \\
  \texttt{serinac@berkeley.edu} \\\And
  Ashton Anderson \\
  Microsoft Research \\
  University of Toronto \\
  \texttt{ashton@cs.toronto.edu} \\\And
  Jake M. Hofman \\
  Microsoft Research \\
  \texttt{jmh@microsoft.com}
}

\begin{document}
\maketitle
\begin{abstract}
With the rapid adoption of LLM-based chatbots, there is a pressing need to evaluate what humans and LLMs can achieve together.
However, standard benchmarks, such as MMLU, measure LLM capabilities in isolation (i.e., ``AI-alone''). 
Here, we design and conduct a user study to \textit{convert} MMLU questions into user-AI conversations, by seeding the user with the question and having them carry out a conversation with the LLM to answer their question.
We release \dataname{}, a new dataset with AI-alone, user-alone, and user-AI data for \numQuestions{} questions and two LLMs, including \numAnswers{} answers and \numConversations{} user-AI conversations.
We find that AI-alone accuracy fails to predict user-AI accuracy, with significant differences across multiple subjects (math, physics, and moral reasoning), and we analyze the user-AI conversations to provide insight into how they diverge from AI-alone benchmarks.
Finally, 
we show that fine-tuning a user simulator on a subset of \dataname{} improves its ability to estimate user-AI accuracies, increasing correlation on held-out questions by more than 20 points, creating possibilities for scaling interactive evaluation.\footnote{Our dataset \dataname{} is available at \huggingface{}.}
\end{abstract}

\input{intro}
\input{related}
\input{study_design}
\input{study_results}
\input{simulator}
\input{conclusion}
\input{limitations}
\input{broader_impacts}

\section*{Acknowledgments}
The authors thank Rich Ciapala for computing support, and Emma Pierson and Joseph Suh for helpful discussions.

\bibliography{custom}

\appendix
\renewcommand\thefigure{\thesection\arabic{figure}}    
\renewcommand\thetable{\thesection\arabic{table}}    

\setcounter{figure}{0}    
\setcounter{table}{0} 
\input{app_study}

\setcounter{figure}{0}    
\setcounter{table}{0} 
\input{app_analysis}

\end{document}

%% file: intro.tex
\section{Introduction}
In 2024, nearly 40\% of US adults reported using generative AI in their everyday lives, an unprecedented rate of adoption for a new technology \cite{bick2024adoption}.
As these models, particularly large language models (LLMs), become more integrated into our lives, it becomes increasingly important to evaluate not only their capabilities in isolation, but also on how effectively they support people in solving a variety of tasks.
However, there is a large gap between human interactions and how standard benchmarks, such as Massive Multitask Language Understanding (MMLU), evaluate models \cite{hendrycks2021mmlu}.
These benchmarks test models on a fixed set of questions, and for each question, they prompt the model with the entire question text and often constrain it to respond with a single multiple choice option as its answer.
In contrast, interactions with human users are far more variable, open-ended, and subject to ambiguity.
Even conditioned on the same underlying intent, users may phrase their prompts differently, leave out information in their early prompts, or rely on context in later prompts. Robust AI models must be designed to handle user interactions in these contexts to provide accurate information and complement human expertise.

Recently, there have been efforts to evaluate how humans interact with LLMs, such as examining real-world conversations using a strong LLM as a judge \cite{lin2024wildbench,li2024arenahard}.
However, these new evaluations have been largely disconnected from standard benchmarks, which are widely used; for example, every LLM released by OpenAI, Google, and Meta, inter alia, has reported its performance on MMLU \cite{openai2023gpt4,google2023gemini,meta2024llama3}.
This disconnect is due to a large distribution shift between benchmark questions and questions asked by real-world users, missing the user's true intent, and missing ground-truth labels to judge the interaction, necessitating techniques like LLM-as-judge.
As a result, it is difficult to directly compare results from standard benchmarks to real-world interactions or to understand how incorporating interactions changes evaluation insights.

Here, we seek to bring these lines of research closer together by directly \textit{converting} benchmarks into user-AI conversations.
We focus on MMLU, as one of the most widely used benchmarks, and design a user study where we seed users with an MMLU question and have them carry out a conversation with an LLM with the intent of answering that question.
For each question, we test the LLM in isolation (i.e., ``AI-alone'') and evaluate the accuracy of a user interacting with the LLM (i.e., ``user-AI''); furthermore, we also gather ``user-alone'' data per question to understand how much users improve with the LLM.
This parallel data has two advantages: first, we can now conduct an apples-to-apples comparison of AI-alone performance, as reported in most papers, vs. user-AI performance on the same questions, so that we can isolate the effects of incorporating interaction into evaluation.
Second, recent works have explored the possibility of \textit{simulating} the user in user-AI conversations \cite{li2024iqaeval} but lack sufficient data for training and testing.
Our approach of ``seeding'' users with a question corresponds naturally to a new way to initialize user simulators, and the large-scale data we collect enables fine-tuning and validating a user simulator on this task, improving the trustworthiness of simulations for AI evaluation. 

Our resulting dataset \dataname{}, which we release publicly, consists of AI-alone, user-alone, and user-AI data for \numQuestions{} questions and two LLMs (GPT-4o and Llama-3.1-8b), with \numAnswers{} answers and \numConversations{} user-AI conversations.
Our study design also includes two user-AI conditions---where the user attempts the question first on their own vs. uses AI directly---to explore nuances in user behavior.
Our study reveals that AI-alone accuracy fails to predict user-AI accuracy, with significant differences across multiple subjects (math, physics, and moral reasoning).
We also analyze the user-AI conversations to understand where user-AI interactions are diverging from AI-alone benchmarks.
Our contributions are summarized as follows:
\begin{itemize}
    \item We design and conduct a user study to convert MMLU questions into user-AI conversations and release a large-scale dataset \dataname{}.
    \item We show that AI-alone accuracy fails to predict user-AI accuracy, across subjects, models, AI-alone methods, and user-AI conditions, and we analyze user-AI conversations to understand where AI-alone and user-AI diverge.
    \item We develop a new user simulator that mimics our user study task and show that fine-tuning our simulator on \dataname{} improves its correlation with real user-AI accuracies by 22-26 points and outperforms baselines.
\end{itemize}
All together, our work helps to reconcile two vital lines of research in AI evaluation, revealing how interactions change evaluation insights and 
paving the way towards scalable interactive evaluation.

%% file: related.tex
\section{Related Work}

\paragraph{Benchmarks.}
In this work, we focus on MMLU as one of the most commonly used LLM benchmarks \cite{hendrycks2021mmlu}.
MMLU is a question-answering (QA) dataset, consisting of multiple choice questions across 57 subjects (which we discuss in detail in Section \ref{sec:study-questions}).
We also draw on the efforts of MMLU-Redux \cite{gema2024mmluredux}, where authors noted some quality concerns in the original MMLU, so they sampled a large number of MMLU questions and manually annotated them for errors.
While we conduct our user study on MMLU, our approach of converting QA benchmarks to a user-AI conversation is general, and could be applied to other QA benchmarks, such as HotPotQA \cite{yang2018hotpotqa} or GSM8K \cite{cobbe2021gsm8k}, as well as adapted to non-QA tasks.

\paragraph{Evaluating human-AI interactions.}
Recently, there have been growing efforts to evaluate AI models based on their interactions with humans.
For example, some works gather real-world interactions (e.g., WildChat \cite{zhao2024wildchat}, ChatbotArena \cite{chiang2024chatbotarena}) and evaluate the interactions (e.g., WildBench \cite{lin2024wildbench}, ArenaHard \cite{li2024arenahard}, MT-Bench \cite{zheng2023llmjudge}, LMSYS-Chat-1M \cite{zheng2024lmsys}), typically using a strong LLM as a judge.
However, as discussed before, it is difficult to directly compare these evaluation results to standard benchmarks, due to the lack of ground-truth user intents and interaction labels, distribution shift in questions, and change in evaluation metric.
Other works have evaluated human-AI interactions in diverse contexts, such as theorem proving \cite{collins2024theorem}, education \cite{google2024learnlm}, co-writing with AI \cite{dhillon2024cowriting}, and collaborating with AI agents \cite{shao2024cogym},  and sought to understand where human-AI combinations outperform either alone \cite{bansal2021whole,vaccaro2024combo}.

Our work builds on \citet{lee2023interaction}, who argue for the need to evaluate human-LM interactions, covering five types of tasks including QA.
Their work includes an exploratory user study where they have users interactively answer MMLU questions; however, they only test 30 questions
and do not explore simulation.
Our study builds on theirs by testing \numQuestions{} questions, at a large enough scale to estimate significant effects and fine-tune a user simulator, and introduces an AI-alone method that is a far more competitive baseline for estimating user-AI results.
Furthermore, our study tests more sophisticated LLMs, complex reasoning subjects, and user-AI effects across levels of question difficulty and user-AI conditions.
While different in domain, our work is also related to \citet{li2024mediq}, who convert medical benchmarks into simulated interactions between a patient and an expert.

\paragraph{Simulation with LLMs.}
LLMs have shown promising capabilities to realistically simulate human behaviors, such as responses to surveys and social science experiments \cite{argyle2023samples,horton2023silicus,hwang2023aligning,hewitt2024social,suh2025subpop} or interactions between humans \cite{park2023generative,chang2024network}.
There is also much interest in developing LLM-based user simulators to scale AI evaluation and training \cite{dubois2023alpacafarm,ren2024bases,kong2024platolm,li2024iqaeval}.
However, LLMs can sometimes produce unrealistic simulations of humans, with risks of bias or uniformity \cite{cheng2023marked,cheng2023compost,bisbee2024perils,wang2024flatten}.
Thus, there is a need to rigorously test whether LLM simulators produce realistic outputs and match insights that we would learn from real humans.
Here, we examine a setting with well-defined simulator goals (i.e., does the simulator match user behavior and accuracy in real user-AI conversations) and release a large-scale dataset that enables training and validation of simulators in this setting.

%% file: study_design.tex
\section{User Study Design}
\label{sec:design}

In this section, we discuss our user study design, including the task flow and interface, how we selected questions, and data collection.
We provide additional details in Appendix \ref{sec:app_study}.

\subsection{Task Flow and Interface}
\begin{figure}
    \centering
    \includegraphics[width=0.8\linewidth]{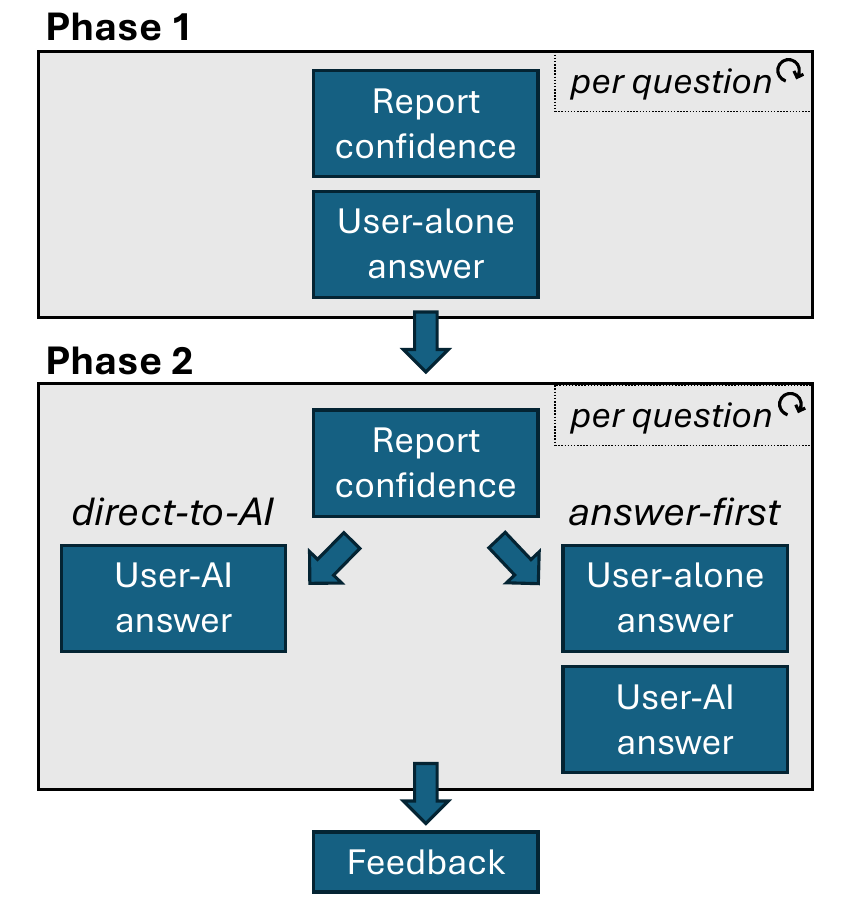}
    \caption{Flow of our user study.}
    \label{fig:flow}
\end{figure}

Figure \ref{fig:flow} shows the flow of our user study.
In Phase 1, users are asked to answer each question to the best of their ability on their own.
In Phase 2, users are asked to chat with an unnamed ``AI Chatbot'' to help them answer their question.
We test two LLMs, contrasting GPT-4o as a strong model and Llama-3.1-8b as a relatively weaker model.
We require interaction in Phase 2---the user cannot move onto the next question without sending a message and we say that low-effort conversations, e.g., only ``hi'', will be flagged---but otherwise, we do not specify at all how the user should interact with the AI Chatbot.
In both phases, users are asked to first report how confident they are about approaching the problem, before attempting to answer it.
This additional question-level variable allows us to analyze how AI assistance helps users across varying levels of confidence.
After Phase 2, all users provide feedback on the task, with free-text responses including whether they found the AI Chatbot helpful and if they saw it make any mistakes. 
In Figure \ref{fig:phase2_screenshot}, we show a screenshot of what users see in Phase 2; in the Appendix, we provide screenshots of all other pages in our task (Figures \ref{fig:consent}-\ref{fig:feedback}).

\begin{figure}
    \centering
    \includegraphics[width=\linewidth]{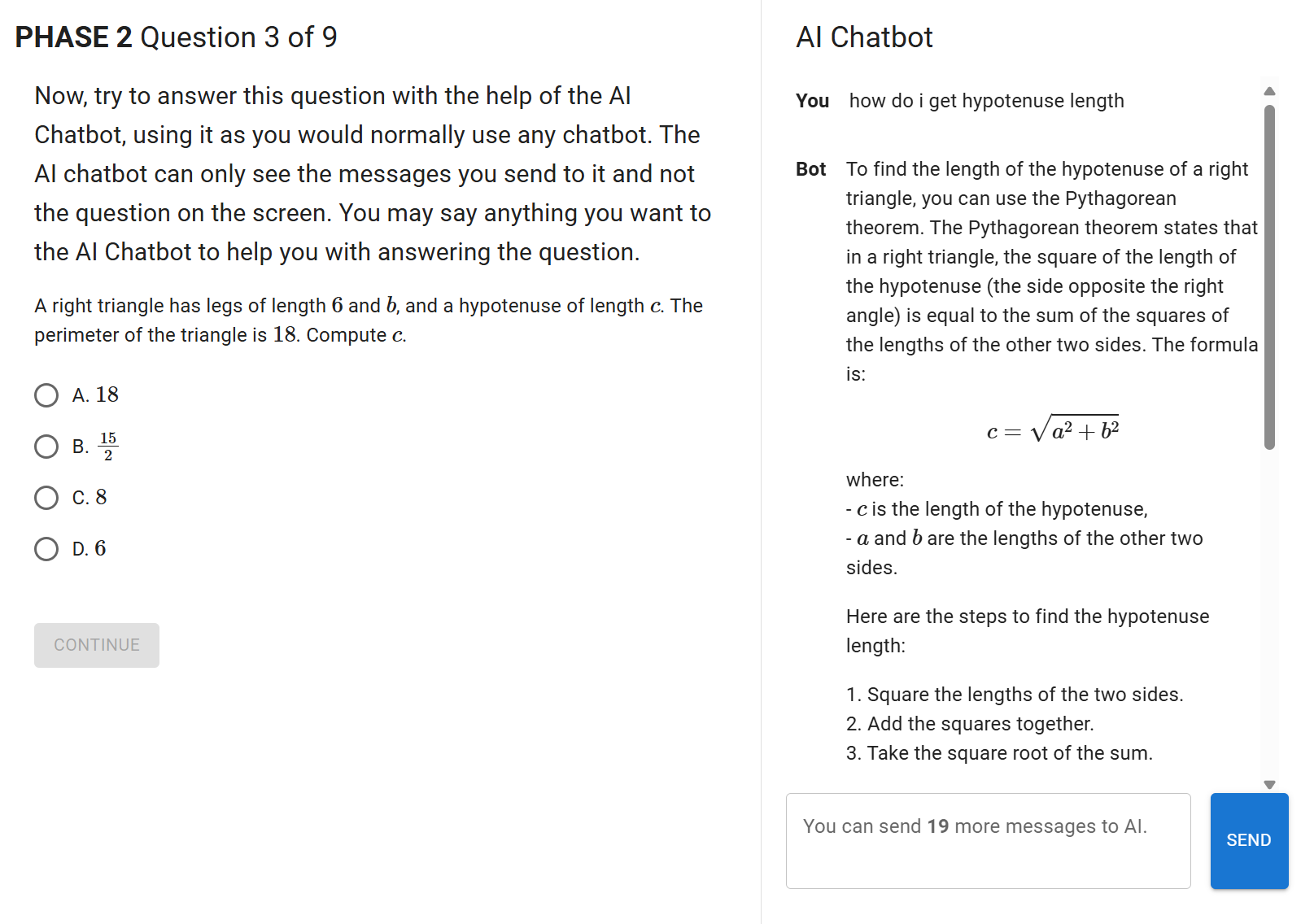}
    \caption{Screenshot from Phase 2 where the user interacts with an AI Chatbot to answer the question.}
    \label{fig:phase2_screenshot}
\end{figure}
\paragraph{Conditions.}
We explore two user-AI conditions: \textit{answer-first} and \textit{direct-to-AI}.
In the \textit{answer-first} condition, the user attempts to answer each Phase 2 question on their own first before answering with AI, but in the \textit{direct-to-AI} condition, they have immediate access to AI for the Phase 2 questions (in both conditions, Phase 1 is all user-alone).
The advantage of \textit{answer-first} is that, for the same question, we can record a user's answer on their own vs. with AI, allowing us to estimate the marginal impact of AI more precisely (i.e., within-subjects), while for \textit{direct-to-AI}, the set of user-alone answers and user-AI answers for a given question come from different users (i.e., between-subjects).
However, we hypothesized that user behavior and accuracy in the user-AI stage could be impacted by the user attempting the answer first, reducing ecological validity if we believe users typically go directly to AI in the real world.
Thus, we keep both conditions, allowing us to test our hypothesis and explore nuances in user behavior.


\paragraph{Incentivization.}
To incentivize participants in our study to answer questions correctly, we included a small bonus of \$0.10 per correct answer, on top of a base compensation of \$5.00 for completing the entire task.
We included these incentives to improve ecological validity, since our study is meant to capture how a real-world user would interact with an AI system if they have a question that they genuinely want to answer.
In Appendix \ref{sec:incentivization}, we discuss pilots we ran with and without incentivization,
as well as how we mitigated risks of cheating with external tools.

\subsection{Question Selection}
\label{sec:study-questions}
\begin{figure}
    \centering
    \includegraphics[width=0.9\linewidth]{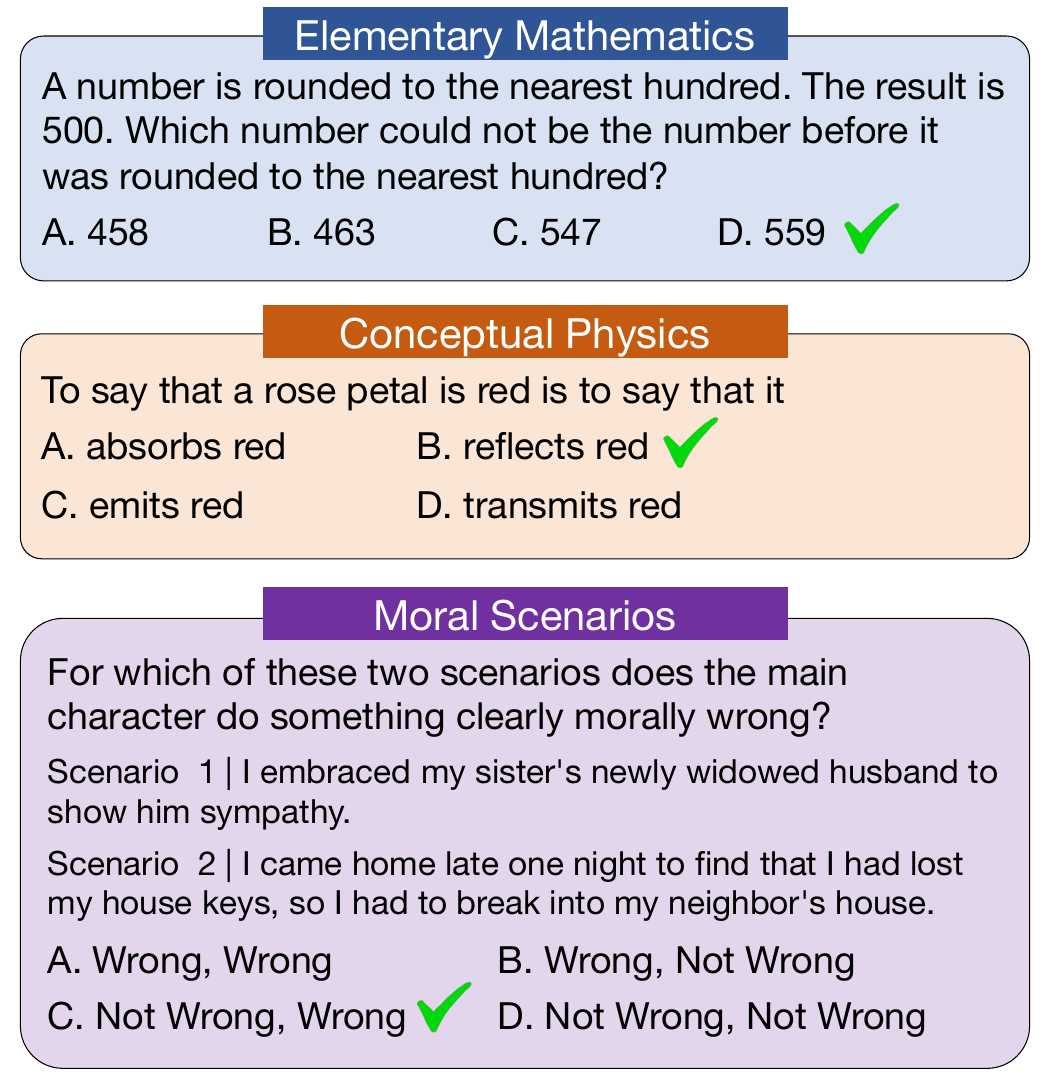}
    \caption{Examples of questions from our user study.}
    \label{fig:question_examples}
\end{figure}
We consider five datasets from MMLU for our experiments: Elementary, High School, and College Mathematics, Conceptual Physics, and Moral Scenarios. 
We include three math datasets since this subject still poses unique challenges for LLMs: for example, the HELM leaderboard \cite{liang2023helm} reports that while GPT-4o's mean accuracy on MMLU is 84\%, its accuracy is only 48\% on High School Math and 51\% on College Math.\footnote{\url{https://crfm.stanford.edu/helm/mmlu/latest/\#/leaderboard}.}
Furthermore, the three math datasets stratify different levels of difficulty for humans, allowing us to explore how user-AI effects change across difficulty levels. 
We also include Conceptual Physics and Moral Scenarios as two other reasoning domains with very different types of problems and differing levels of human expertise.
In Figure \ref{fig:question_examples}, we provide examples of questions from these datasets, showcasing their diversity.

To aid with question selection, we use the annotations from MMLU-Redux \cite{gema2024mmluredux}.
The authors recognized occasional quality issues with the original MMLU, so for each MMLU dataset, they sampled 100 questions from the test set uniformly at random and labeled them for errors. 
While they found many errors in some datasets (e.g., Virology), the majority of the questions (92\%-99\%) in the datasets we chose passed their review.
As a second layer of quality control, we also ran OpenAI's advanced reasoning o1 model over the 100 questions and manually checked the questions that o1 did not get correct.
We kept the intersection of questions that passed MMLU-Redux's inspection and ours (with o1's help).

\paragraph{Batches.} 
To reduce variance in the number of answers that each question received, we organized the questions into batches and selected a random batch per user, instead of selecting each question randomly.
For the math questions, each batch consisted of 5 elementary, 5 high school, and 2 college questions.
We included fewer college questions since we found in pilots that college questions were too difficult for most users, so they tended to defer to the LLM's first answer without much interaction.
Based on the number of questions that passed inspection, we were able to create 19 math batches, with 95 elementary, 95 high school, and 38 college questions in total.
For Conceptual Physics and Moral Scenarios, we constructed 7 batches of size 12, resulting in 84 questions for each subject.

\subsection{Data Collection}
We recruited workers on Prolific to participate as users in our study (see eligibility criteria in Appendix \ref{sec:app_study}).
For our full pre-registered study, we recruited 650 workers, and we also ran two medium-sized pilots (100 workers without incentives and 60 workers with incentives).
When a user began the study, they were randomly assigned to one of the three subjects (60\% probability for math, 20\% for conceptual physics, and 20\% for moral scenarios) and assigned uniformly at random to one of that subject's question batches, one of the two user-AI conditions, and one of the two models (GPT-4o and Llama-3.1-8b).
Within the question batch, 3 questions were randomly assigned to Phase 1 and 9 to Phase 2. 
We also included an attention-check question for every user, which we found the vast majority (over 99\%) of users passed. 

Compiling data over the three runs, we have 10,828 confidence answers, 7,148 user-alone answers, and \numConversations{} user-AI answers and conversations in \dataname{} (see Table \ref{tab:data_stats} for additional data statistics).
While we include data from all three runs in \dataname{} to provide a larger resource for the community, for our analyses in the rest of the paper, we only use data from the workers in our full pre-registered study so that populations within our analysis are entirely comparable. 

%% file: study_results.tex
\section{Experimental Results}
\label{sec:study-results}
\begin{figure*}
    \centering
    \includegraphics[width=\linewidth]{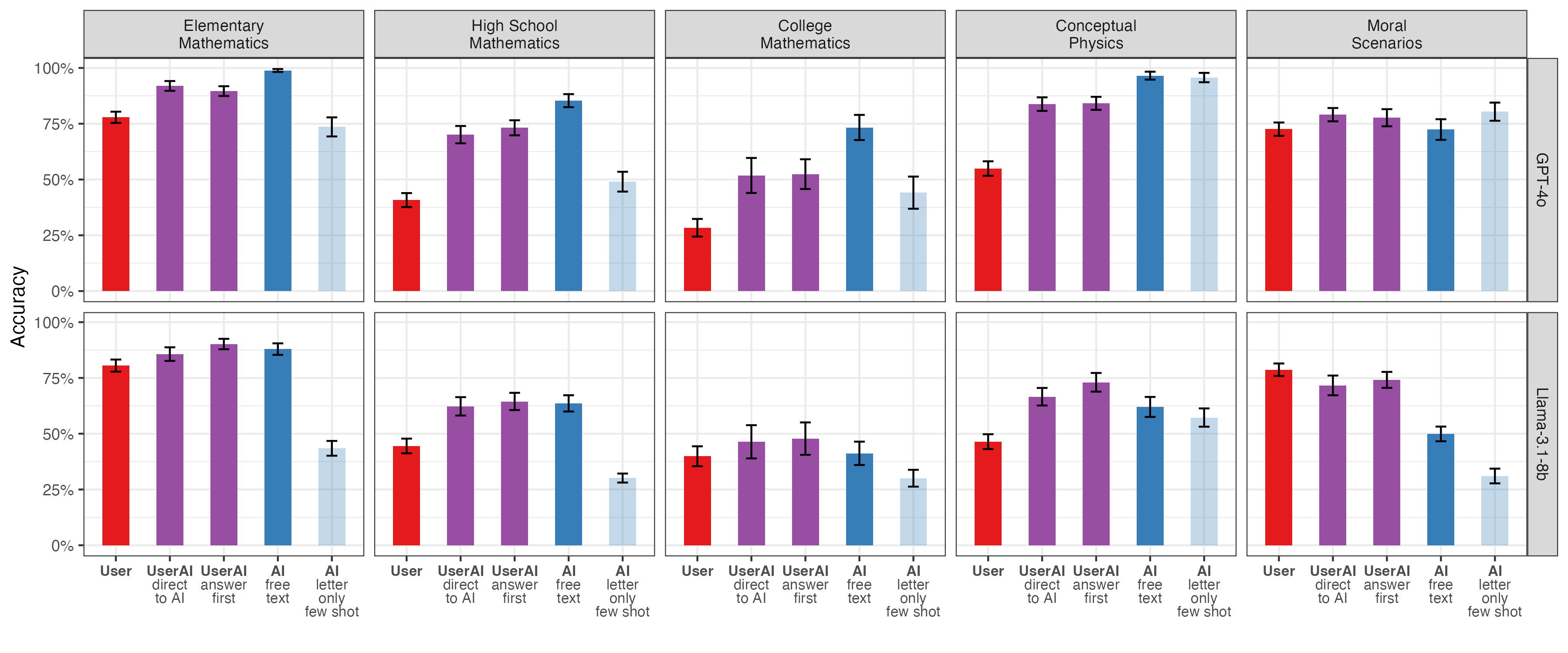}
    \caption{Mean accuracy per model and dataset, comparing user-alone (red), user-AI (purple), AI-alone free-text (dark blue), and AI-alone letter-only few-shot (light blue). See Tables \ref{tab:gpt4o-results}-\ref{tab:llama-results} for numbers and statistical tests.}
    \label{fig:mean_accs}
\end{figure*}

In this section, we describe our experimental results, including how we conducted AI-alone experiments, comparisons of AI-alone vs. user-AI results, and analyses of the user-AI conversations.
For our main results comparing AI-alone vs. user-AI, we preregistered our analyses on AsPredicted.\footnote{\url{https://aspredicted.org/n84n-sn3f.pdf}.} 
We provide additional results and methodological details (e.g., statistical tests) in Appendix \ref{sec:app_analyses}. 

\subsection{AI-Alone Experiments}
\label{sec:ai_alone}
Our goal in this work is to understand how evaluation conclusions change when we move from AI-alone to user-AI settings.
However, even for a fixed benchmark, there can be multiple ways to evaluate an LLM on its own.
First, we try \textit{letter-only} methods, which require the model to answer with only a single letter corresponding to the selected answer option (``A'' through ``D'').
This is the method used by \citet{lee2023interaction}, along with various leaderboards, such as HELM \cite{liang2023helm}, to standardize the answer format.
We try two letter-only variants, zero-shot and few-shot, where we prepend the 5 examples from the MMLU ``dev'' set to the prompt as in-context examples. 

We also introduce a more realistic AI-alone technique which serves as a better proxy for user experience by not constraining the model's response format.
The method, which we call \textit{free-text}, is very simple: (1) prompt the evaluated model with the concatenated question text and answer options, without any additional instructions, (2) use GPT-4o to extract an answer (if any) from the response.
We include the full prompts for all three AI-alone methods in Listings \ref{lst:few_shot_prompt}-\ref{lst:free_text_prompt_2}.

We ran these three AI-alone methods on the two models and all \numQuestions{} questions from our user study, gathering 50 answers per model and question.
As shown in Figure \ref{fig:mean_accs}, our few-shot letter-only results for GPT-4o approximately match those reported on the HELM leaderboard per dataset (which is also few-shot letter-only, but uses the entire MMLU test sets).
While prior work, like HELM, often uses temperatures of 0 for multiple choice QA, we used a temperature of 0.7, since we wanted to perfectly match the model parameters used in the user study, and 0.7 is a more realistic temperature for real-world AI chatbots. 

\subsection{AI-Alone vs. User-AI}
\label{sec:ai_alone_vs_user_ai}
\paragraph{Dataset-level accuracy.}
We visualize our main results in Figure \ref{fig:mean_accs}, which shows mean accuracy per model and dataset, over user-alone (red), user-AI (purple), and AI-alone (blue).
First, we see that few-shot letter-only (light blue) is a very poor predictor of user-AI performance, with a mean absolute deviation of 21 percentage points, averaged over the 10 dataset and model pairs.
With a few exceptions---specifically Conceptual Physics for Llama-3.1-8b and College Mathematics and Moral Scenarios for GPT-4o---all differences are statistically significant.
Results are similar for zero-shot letter-only, which we report in Tables \ref{tab:gpt4o-results}-\ref{tab:llama-results}.
Notably, our AI-alone method, free-text (dark blue), is a much better predictor of user-AI accuracy, reducing the mean absolute deviation to 10 percentage points.
However, it still differs significantly from user-AI performance, notably for Moral Scenarios with Llama-3.1-8b and for all datasets except Moral Scenarios with GPT-4o.

Our results also reveal the complexity of combining humans and AI, as the size of gaps and  ordering between user-alone, user-AI, and AI-alone vary over models and datasets.
For example, for the math datasets, GPT-4o performs quite well on its own (using free-text), while humans struggle on their own, especially for high school and college. 
In these cases, user-AI accuracy is between the two, significantly better than user-alone and significantly worse than AI-alone.
Meanwhile, Llama-3.1-8b performs significantly worse than GPT-4o on the math datasets, but we do not see a further drop in performance from AI-alone to user-AI.
In the following section, we uncover counterveilling factors that explain these results: on one hand, users introduce ambiguity compared to AI-alone methods, which include the entire question text and answer options; on the other hand, users can sometimes recognize mistakes in AI reasoning, of which there are more for Llama-3.1-8b.
Finally, our results reveal that even when AI-alone benchmarks report a large gap in performance between two models, this gap can become much smaller after incorporating user interactions.
Comparing GPT-4o and Llama-3.1-8b, their average gap in AI-alone free-text accuracy is
25 percentage points, but this gap shrinks to less than 10 percentage points in user-AI interactions (9 percentage points for \textit{direct-to-AI} and 5 percentage points for \textit{answer-first}). 

\paragraph{Question-level accuracy.}
Besides mean accuracy, we can also measure the correlation in per-question accuracies.
We find that the Pearson correlation between AI-alone free-text and user-AI is only $r=0.45$ for \textit{direct-to-AI} and $r=0.46$ for \textit{answer-first}.
While correlations may be lower because per-question user-AI accuracies are imperfectly measured, the free-text correlation is still well below what we would expect if user-AI accuracies were drawn from the same distribution as free-text, which would range from $r=0.88$ to $0.94$ (Section \ref{sec:statistical_tests}).

We also examine the correlation with per-question \textit{differences} in user-AI and user-alone accuracy, since it may be more reasonable to expect AI-alone to predict the improvement the user makes with AI assistance, instead of the overall accuracy.
However, the correlations remain low, at $r=0.26$ for \textit{direct-to-AI} and $r=0.27$ for \textit{answer-first}, showing that AI-alone cannot predict improvements well either. 
Finally, we fit a linear model to try predicting a question's user-AI accuracy from its user-alone and AI-alone accuracies.
The fitted model yields a correlation of 0.55 for predicting \textit{answer-first} accuracies and 0.63 for predicting \textit{direct-to-AI} accuracies, demonstrating that user-AI accuracy also cannot be reliably predicted from user-alone and AI-alone accuracy.

\subsection{Characterizing User-AI Conversations}
\label{sec:conversations}
Our summary results show that user-AI accuracies are significantly different from AI-alone accuracies. To better understand what drives these differences we use a separate LLM as an annotator to characterize the user-AI conversations. For each user-AI conversation, we gather the full log of the conversation and the correct answer to the question, and prompt a separate instance of GPT-4o to use this information to extract the answers to several classification questions: 1) whether the first user question was a near-exact rephrasing of the original question to which an intelligent person or AI would respond with the correct answer among the answer choices; 2) whether the first AI answer is correct; 3) whether the AI provides an answer to the question at hand more than once; and 4) which of the four answer choices the AI's first answer corresponds to, or ``none'' if there is no such answer (Listing \ref{lst:conversation_prompt}). 

\begin{figure}
    \centering
    \includegraphics[width=\linewidth]{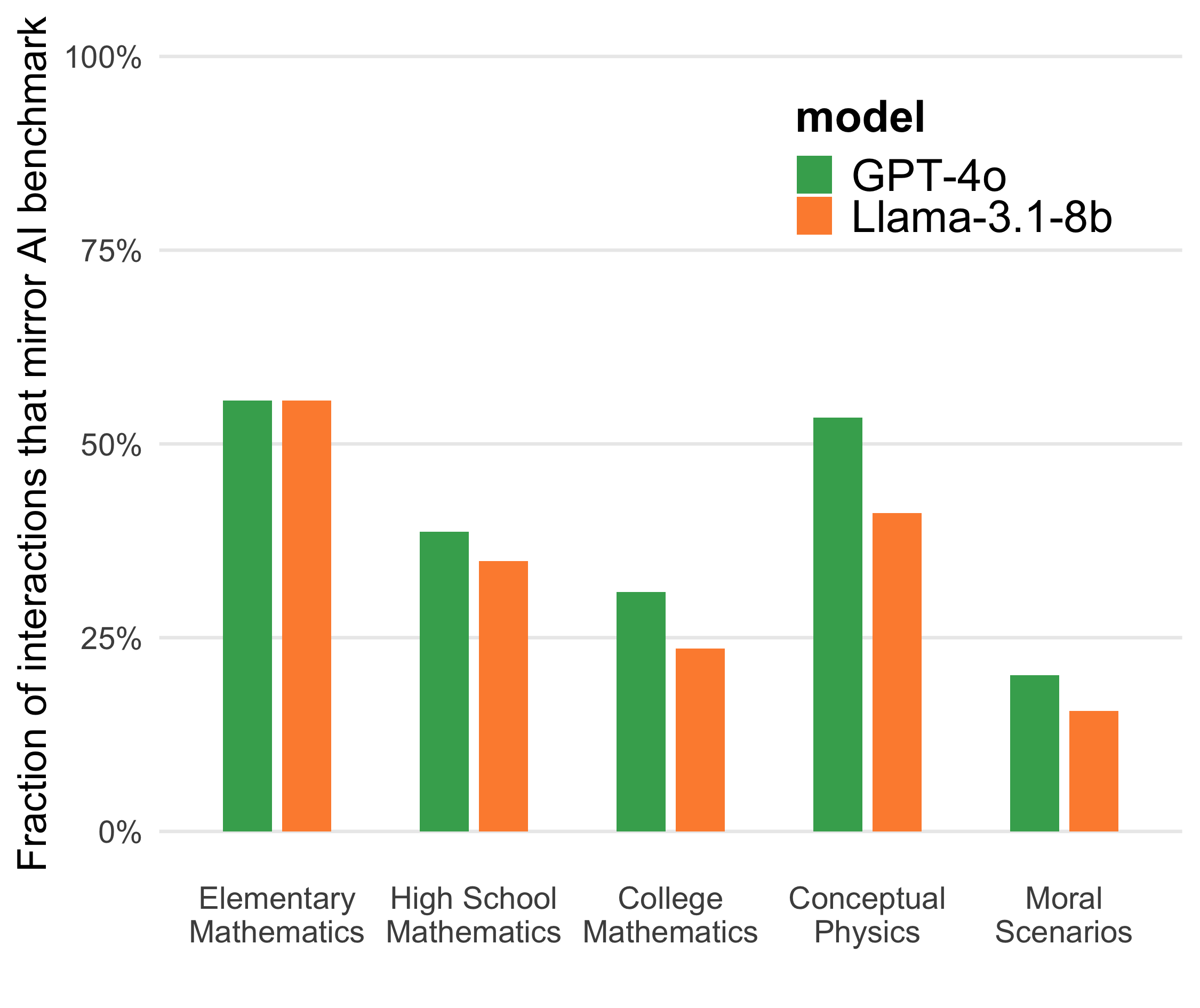}
    \caption{Fraction of user-AI interactions that mirror AI benchmark, by subject and model.}
    \label{fig:mirror_ai}
\end{figure}
How often does the conversation follow what we might expect if AI benchmarks were faithful proxies of human-AI interaction? We say a conversation \emph{mirrors} an AI benchmark if (1) the user's first substantive prompt is a near-exact rephrasing of the benchmark question (otherwise the user is injecting their own knowledge or information into the interaction), (2) the LLM responds with only a single answer during the entire interaction, and (3) the user submits that answer. 
In Figure~\ref{fig:mirror_ai}, we see that only $39.8\%$ of all interactions mirror AI benchmarks, revealing the extent to which user-AI interactions diverge from AI benchmarks.

Using data from the \textit{answer-first} condition also reveals that AI helps humans more often than it hinders them. 
When the same user answers a question first without AI and then with AI assistance, more than half (54\%) of incorrect user-alone answers are corrected with AI support, while only 10\% of correct user-alone answers turn incorrect with AI assistance.
This data also allows us to look more closely at how often user-AI interactions improve on AI-alone performance.
Compared to AI-alone free-text, user-AI accuracy is lower for 57\% and higher for 15\% of questions for GPT-4o, and lower for 39\% and higher for 44\% of questions for Llama-3.1-8b (keeping questions where we have at least 5 user-AI answers for the AI system). 
Thus, we see effects in both directions, and there are certainly cases where user-AI improves on AI-alone, especially for the weaker model.
Below, we analyze both types of cases in more detail.

\paragraph{Cases where interaction introduces errors.}
First, we study cases where user interaction introduces errors, by focusing on questions where AI-alone free-text is always correct (accuracy of 100\% over 50 trials) but the user-AI interaction results in the wrong answer.
This could happen either if the user decided to ignore the AI's correct answer (e.g., if they believed they knew the answer or due to lack of effort) or if the change from AI-alone prompting to user prompting resulted in the AI no longer providing the correct answer. 
We find much more evidence for the latter.
Among over 300 of these interactions with GPT-4o, the model only provided the correct answer in 26\% of interactions, and in the remaining interactions, the model either did not provide any clear answer (67\%) or provided a wrong answer (7\%).
Among 122 interactions with Llama-3.1-8b (there are fewer because there are fewer questions where Llama-3.1-8b achieves perfect accuracy on its own), the model only provided the correct answer in 20\% of interactions, instead providing no clear answer in 68\% of interactions and the wrong answer in 12\% of interactions. 

We also find that in the majority of these interactions (66.2\% for GPT-4o and 59.8\% for Llama-3.1-8b), the user's first substantive prompt is \textit{not} a near-exact rephrasing of the benchmark question, providing further evidence for our hypothesis that the change in accuracy is largely due to the shift in prompting from AI-alone benchmarks to human users.
We find that a primary source of divergence is the user asking a related but different question, which is often ambiguous (e.g., leaving out critical information for a math problem). 

\paragraph{Cases where interaction corrects AI errors.}
Next, we study the opposite scenario: questions where AI-alone free-text's accuracy is poor (below 10\% over 50 trials) but the user-AI interaction arrives at the correct answer (there are approximately 100 such interactions for each model). 
In a substantial fraction of these interactions, the AI model actually provided the correct answer to the user (40\% for GPT-4o and 32\% for Llama-3.1-8b), suggesting that the user's prompting enabled the AI to arrive at the correct answer, even though it could not on its own.
Notably, we find that in around 10\% of these interactions the AI's responses shift from incorrect or unclear to correct over the course of the conversation, highlighting how user prompting can elicit better answers and underscoring the importance of multi-turn analysis beyond static benchmarks.

Even when the AI is not able to arrive at the correct answer, we find that users are sometimes still able to correct the mistake and select the right answer.
We visualize the rates of these corrections in Figure \ref{fig:user_corrects_ai}. 

%% file: simulator.tex
\section{Simulating User-AI Conversations}
\label{sec:simulator}
From our user study, we showed that incorporating user interactions significantly changes evaluation conclusions, compared to AI-alone evaluation. 
However, data from human users is costly and time-consuming to collect, motivating the development of a user simulator to scale interactive evaluation.
In this section, we describe our user simulator and present experimental results.

\subsection{Fine-Tuning a User Simulator}
\begin{figure}
    \centering
    \includegraphics[width=\linewidth]{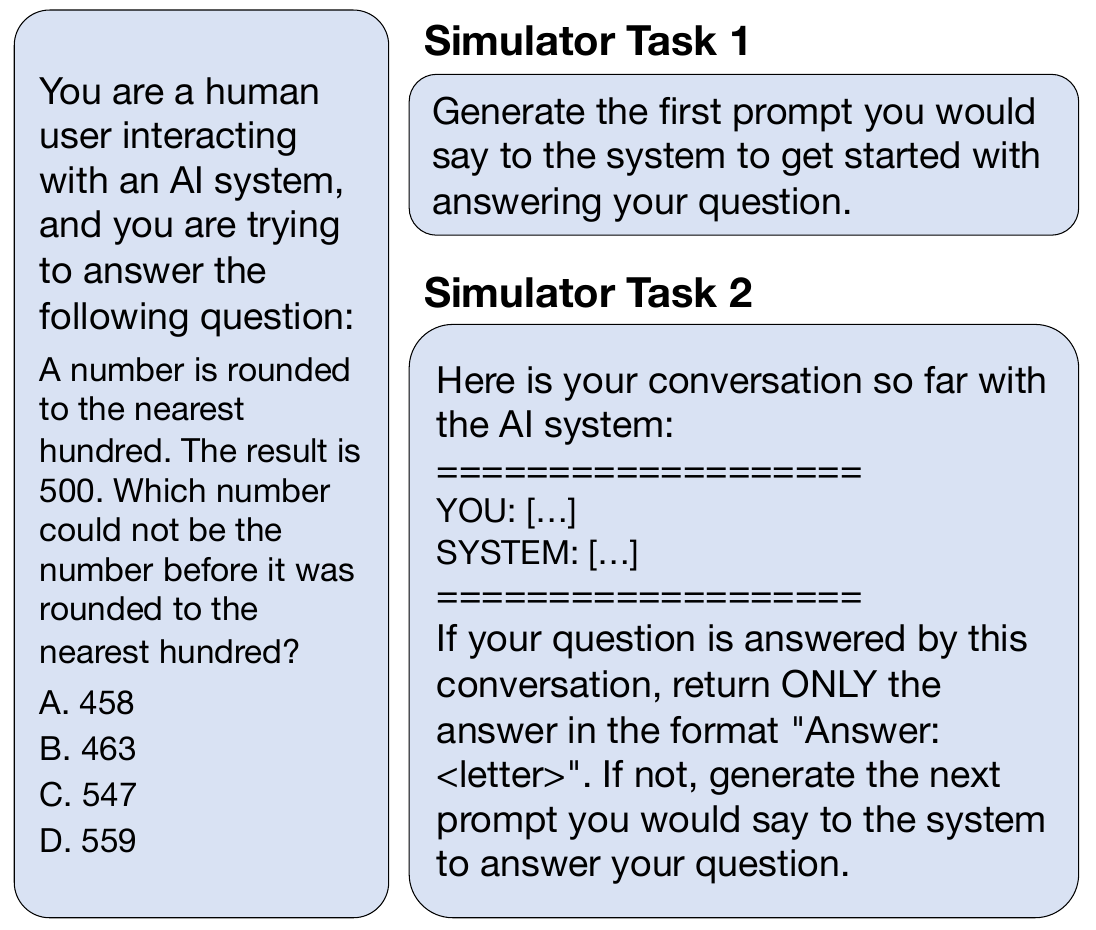}
    \caption{Example of prompts to our two-step user simulator, using one of the example questions from Figure \ref{fig:question_examples}. See Listings \ref{lst:two_step_sys_prompt}-\ref{lst:two_step_user_prompt_2} for complete prompts.}
    \label{fig:simulator}
\end{figure}
\begin{table*}[]
    \centering
    \small
    \begin{tabular}{c|c|c|c|c|c||c|c|c|c}
        & & \multicolumn{4}{c}{\textbf{AI: GPT-4o}} & \multicolumn{4}{c}{\textbf{AI: Llama-3.1-8b}} \\
         Type & Method & Corr. $\uparrow$ & MAE $\downarrow$ & BLEU $\uparrow$ & ROUGE $\uparrow$ & Corr. & MAE & BLEU & ROUGE \\
        \hline 
        AI-alone & Letter-only few-shot & 0.30 & 0.31 & -- & -- & 0.21 & 0.40 & -- & -- \\
        AI-alone & Free-text & 0.49 & 0.20 & -- & -- & 0.61 & 0.20 & -- & -- \\
        Sim-AI & IQA-EVAL & 0.50 & 0.18 & 0.085 & 0.311 & 0.43 & 0.22 & 0.086 & 0.313 \\
        Sim-AI & Two-Step & 0.41 & 0.19 & 0.102 & 0.347 & 0.39 & 0.23 & 0.102 & 0.346 \\
        Sim-AI & \dataname{}-Sim & \textbf{0.63} & \textbf{0.15} & \textbf{0.261} & \textbf{0.460} & \textbf{0.65} & \textbf{0.17} & \textbf{0.258} & \textbf{0.457} \\
        \hline 
    \end{tabular}
    \caption{Comparing to user-AI conversations: AI-alone methods, IQA-EVAL \cite{li2024iqaeval}, and the two-step simulator before (Two-Step) and after fine-tuning on \dataname{} (\dataname{}-Sim). Top-performing is bolded.}
    \label{tab:simulation_results}
\end{table*}

We define a new user simulator that we can fine-tune on our collected user data, by mimicking the experience of users in our study.
First, we seed the user simulator with the MMLU question, as we did with human users in our study, and we tell the simulator to interact with an AI system to answer its question (Figure \ref{fig:simulator}, left).
Then, we break the simulator's task into two subtasks: (1) when there is no conversation yet, we prompt the simulator to generate its first prompt as a user (Figure \ref{fig:simulator}, top right), (2) given the conversation so far, we prompt the simulator to either answer the question in the form ``Answer: LETTER'', if the question has been answered by the conversation, or if not, generate the next user prompt (Figure \ref{fig:simulator}, bottom right).

We then transform the real user-AI conversations from our study into training examples for supervised fine-tuning.
Each conversation with $k$ user utterances yields $k+1$ training examples: one example in the Task 1 format where the gold standard response is the real user's first utterance; $k-1$ examples in the Task 2 format where the gold standard response is each of the remaining utterances (providing the conversation up to that utterance); and one example in the Task 2 format with the full conversation and the gold standard response being ``Answer: LETTER'' corresponding to the user's selected multiple choice option. 

\subsection{User Simulator Experiments}
For these experiments, we use GPT-4o as our simulator.
We try four baselines: the two AI-alone methods, the two-step simulator without fine-tuning, and the user simulator from IQA-EVAL \cite{li2024iqaeval}.
Their simulator, designed with prompt engineering, receives a prompt consisting of a role description (``You are mimicking a human.''), a task description (``You are trying to choose the correct answer for the given question.''), and discussion instructions (e.g., ``In each turn, please only ask one sub-question to interact with the assistant.''); see Listing \ref{lst:iqa_eval_prompt} for the full prompt.
We compare these baselines to our model, the two-step simulator fine-tuned on \dataname{} (``\dataname{}-Sim'').

In our fine-tuning experiments, we randomly split the questions from our user study into 60\% for training ($n=237$) and withheld 40\% for testing ($n=159$), and we fine-tuned on all user-AI conversations for the train questions.
For all three simulator methods, we test them on the held-out test questions by generating conversations entirely from scratch, given only the question
(in contrast, an easier but less realistic set-up would be to provide the real conversation up to the $n^{th}$ turn and have the simulator generate the next user utterance).

\paragraph{Evaluation metrics.}
We generate 10 simulator-AI conversations per test question and compare to real user-AI conversations for the same question and AI system.
To evaluate whether accuracies are similar, we measure the correlation and mean absolute error (MAE) between simulator-AI vs. user-AI accuracies, keeping test questions where we have at least 5 user-AI answers ($n=132$ and $n=124$ for GPT-4o and Llama-3.1-8b, respectively).
To evaluate whether the simulator's generated utterances are realistic, we measure the average BLEU and ROUGE scores of the simulator's first prompt compared to the real user's first prompt.

\paragraph{Results.}
As shown in Table \ref{tab:simulation_results}, fine-tuning our simulator yields large gains, with a 22-26 point increase in correlation and a 21-26\% decrease in MAE.
As shown in Figures \ref{fig:gpt4o-sim}-\ref{fig:llama-sim}, a primary failure mode of the simulator before fine-tuning is that it cannot replicate human mistakes and greatly overestimates user-AI performance, producing far more questions with accuracies of 100\% than we see in the real user-AI distribution, while the fine-tuned simulator matches the real distribution more closely.
We also find that fine-tuning improves the realism of the simulator's generated utterances, with 11-16 point improvements in BLEU and ROUGE.
The fine-tuned simulator also outperforms both AI-alone methods and IQA-EVAL across metrics.

%% file: conclusion.tex
\section{Conclusion}
We have shown that evaluation conclusions change significantly from AI-alone benchmarks to user-AI interactions, across question domains, AI models, AI-alone methods, and user-AI conditions.
Our results motivate the need for more realistic evaluations of AI models that incorporate user interactions.
However, this goal is difficult to achieve, as user data is expensive to collect.
To make this goal more feasible, we both release a new large-scale dataset of user interactions, \dataname{}, and demonstrate the potential of building user simulators to scale interactive evaluation.

The changes we see from AI-alone to user-AI accuracies are often large enough to affect qualitative conclusions about the models.
For example, what can seem like a large disparity between models on AI-alone benchmarks (e.g., 25 percentage point gap between GPT-4o and Llama-3.1-8b on free-text) can shrink to much smaller gaps after incorporating user interactions (e.g., 5 point gap for \textit{answer-first}). 
These changes could impact real-world decisions, such as which model to deploy (e.g., a lightweight, on-device model that performs only slightly worse than a much larger off-device model might be preferable in some circumstances). 

To this end, in future work we hope to understand how AI-alone benchmarks are currently used to make decisions \cite{hardy2024marketing} and how those decisions might change after taking into account human interactions.
We also hope to expand our analysis to more benchmarks and non-QA tasks.
Finally, we hope to develop training techniques to build even more realistic user simulators: while we see large gains from fine-tuning on \dataname{}, the best correlations only reach 0.63, leaving room for future improvement and innovation.


%% file: limitations.tex
\section{Limitations}
Our work has several limitations, which we tried to mitigate but should be taken into consideration when interpreting the results.

\paragraph{Coverage.} 
Our user study has limited coverage of possible benchmarks and user tasks. We chose to focus on the MMLU benchmark \cite{hendrycks2021mmlu} and question-answering as our task, since MMLU is one of the most popular LLM benchmarks and it covers a wide range of subjects, so we could test multiple subjects in comparable ways and with minimal changes to our user study. We began with question-answering since we can naturally transform a benchmark question into a user-AI conversation, where the user is trying to answer the question. However, future work should investigate whether results are consistent on other benchmarks and/or tasks, especially more open-ended generation tasks that are common in real-world user-AI interactions \cite{zhao2024wildchat}.

\paragraph{Ecological validity.}
Our user study is meant to capture how a user would act if they have a question in mind and they are interacting with an AI system to answer their question.
However, since we wanted to match the user's underlying question with the MMLU questions, we had to tell the user what question to answer, which could lead to different behavior compared to if they were intrinsically motivated to answer a question.
To mitigate this, we included a small incentive (\$0.10 per correct answer), so that they would try to get the correct answer, and we filtered out users who failed the attention check; however, it is still possible that users' behaviors would be different in the real world.
Our study setting was also different from real world question-answering: we recruited workers on Prolific to do our study, where they answered 13 questions consecutively in our interface.
Still, we tried to match real-world settings, such as choosing models they might interact with in the real world (e.g., GPT-4o), using realistic model parameters (e.g., temperature of 0.7), and not guiding their prompts to the AI system at all, besides requiring at least one interaction per question. 

%% file: broader_impacts.tex
\section{Broader Impacts and Ethical Considerations}
Our work is driven by broader impacts: we seek to make AI evaluation more realistic and human-centered, by investigating how evaluation conclusions change when we incorporate human interactions.
With our user study, we show that evaluation conclusions change significantly from AI-alone to user-AI settings (for the same set of questions), and these results hold over different subject areas, AI models, AI-alone methods, and user-AI conditions.
We hope that our work motivates AI researchers and practitioners to think more carefully about human-AI interactions when they evaluate AI systems, instead of only using AI-alone benchmarks.

The direction of evaluating human-AI interactions also raises some ethical considerations.
First, we should seek to recruit diverse human participants, since an AI system that works well for one individual or group may not work well for another (e.g., depending on ability, language, preferences, etc.).
Second, user studies should be run ethically: participants should be paid fairly, they should provide informed consent about how their data will be used, and their data should be anonymized and personal information removed (e.g., if they tell the AI system their name).
Third, the possibility of simulating humans in human-AI interactions is exciting and could make interactive evaluation feasible at scale, but LLM-based simulations of humans also have risks that need to be addressed, such as their possibilities for stereotyping, bias, and flattening populations \cite{cheng2023compost,cheng2023marked,wang2024flatten}.
Researchers hoping to build and deploy user simulators should extensively probe for such biases, especially if user demographics are provided in simulator prompts.

%% file: app_study.tex
\section{Details on User Study}
\label{sec:app_study}

\subsection{Task Details}
All studies were reviewed and approved by the Microsoft Research Institutional Review Board (IRB \#10999) and informed consent was obtained from all participants prior to participation.

\paragraph{Recruitment.}
We recruited workers on Prolific to participate in our study.
All Prolific workers who were located in the US, fluent in English, and had \textit{not} participated in one of our pilots were eligible for our study.
We used Prolific's standard sample, which distributed our study to available participants. 
Based on early pilots, we estimated that the task took around 25 minutes.
We paid all participants \$5.00 upon completion of the entire task.
We experimented with offering a small bonus per correct answer, which we discuss in Section~\ref{sec:incentivization}.

\paragraph{Participant demographics.}
The demographic data we received from Prolific about the workers who participated in our study included their country of birth and residence, nationality, fluent languages, age, sex, ethnicity, and employment status.
As a result of our eligibility criteria, all workers resided in the US and were fluent in English.
We also found that the vast majority were born in the US (95\% of the workers who participated in our full study), listed the US as their nationality (99\%), and listed English as their primary language (99\%).

\begin{figure*}
    \centering
    \includegraphics[width=\linewidth]{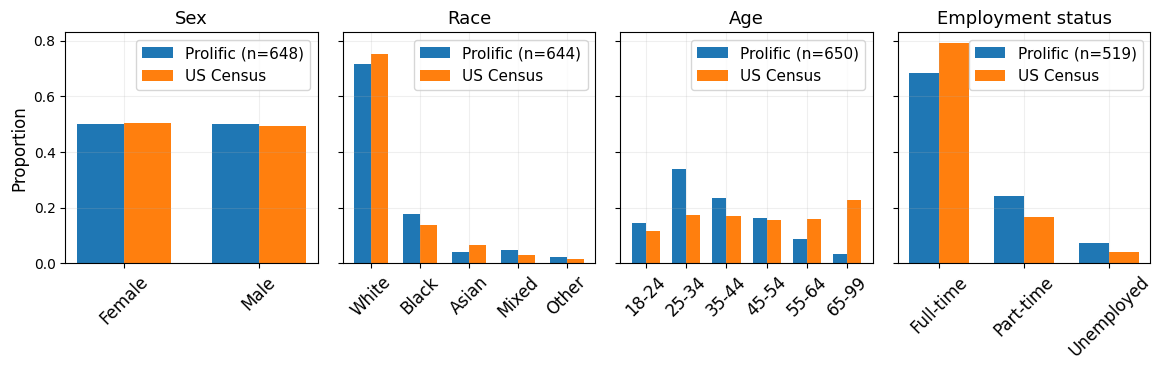}
    \caption{Comparing the marginal distributions of sex, race, age, and employment status for Prolific workers (blue) vs US Census (orange). We only keep data in categories that appear in both datasets (e.g., we drop data values of \texttt{DATA\_EXPIRED} or ``Prefer not to say'' in Prolific), which is why we sometimes have fewer than 650 Prolific workers (see $n$ in the legends). We then normalize the distributions over the kept categories to sum to 1.}
    \label{fig:dist_comparison}
\end{figure*}
Since the workers in our study were primarily from the US, we compare the other demographic variables to the demographic distribution of the US population.
We acquire the joint distribution of sex, race, and age from the \citet{census2024quickfacts}, using their Monthly Postcensal Resident Population
for July 2023 (the latest month before projections). 
For employment status, we use statistics from the \citet{bureau2025labor}, using their estimates for February 2025.

In Figure~\ref{fig:dist_comparison}, we compare each marginal distribution over the Prolific workers vs. the US population. 
The sex distributions are very well-matched: the proportions for "female" and "male" are (0.50, 0.50) on Prolific and (0.51, 0.49) in the US Census. The race distributions are also decently well-matched: using the categories provided to us by Prolific, the proportions for “White”, “Black”, “Asian”, “Mixed”, and “Other” are (0.72, 0.18, 0.04, 0.05, 0.02) on Prolific and (0.75, 0.14, 0.06, 0.03, 0.02) in the US Census. The employment status distribution is slightly different: the proportions for “Full-Time”, “Part-Time”, and “Unemployed (and job seeking)” are (0.68, 0.24, 0.07) on Prolific and (0.79, 0.17, 0.04) in US Labor Statistics, so individuals working full-time are slightly underrepresented while others are overrepresented. We see the largest divergence between distributions on age: individuals aged 25-34 are very overrepresented (0.34 vs. 0.17) while individuals aged 65 and above are very underrepresented (0.03 vs. 0.23). 
These trends are expected, given that younger individuals and those who are not employed full-time are likelier to work on crowdsourcing platforms \cite{ipeirotis2010turk,posch2022crowd}; the younger lean of Prolific workers may also match the younger lean of generative AI users \cite{bick2024adoption}.
Even so, the non-representative nature of workers on crowdsourcing platforms should be taken into account as a limitation when interpreting our results.

\begin{figure*}
    \centering
    \includegraphics[width=0.8\linewidth]{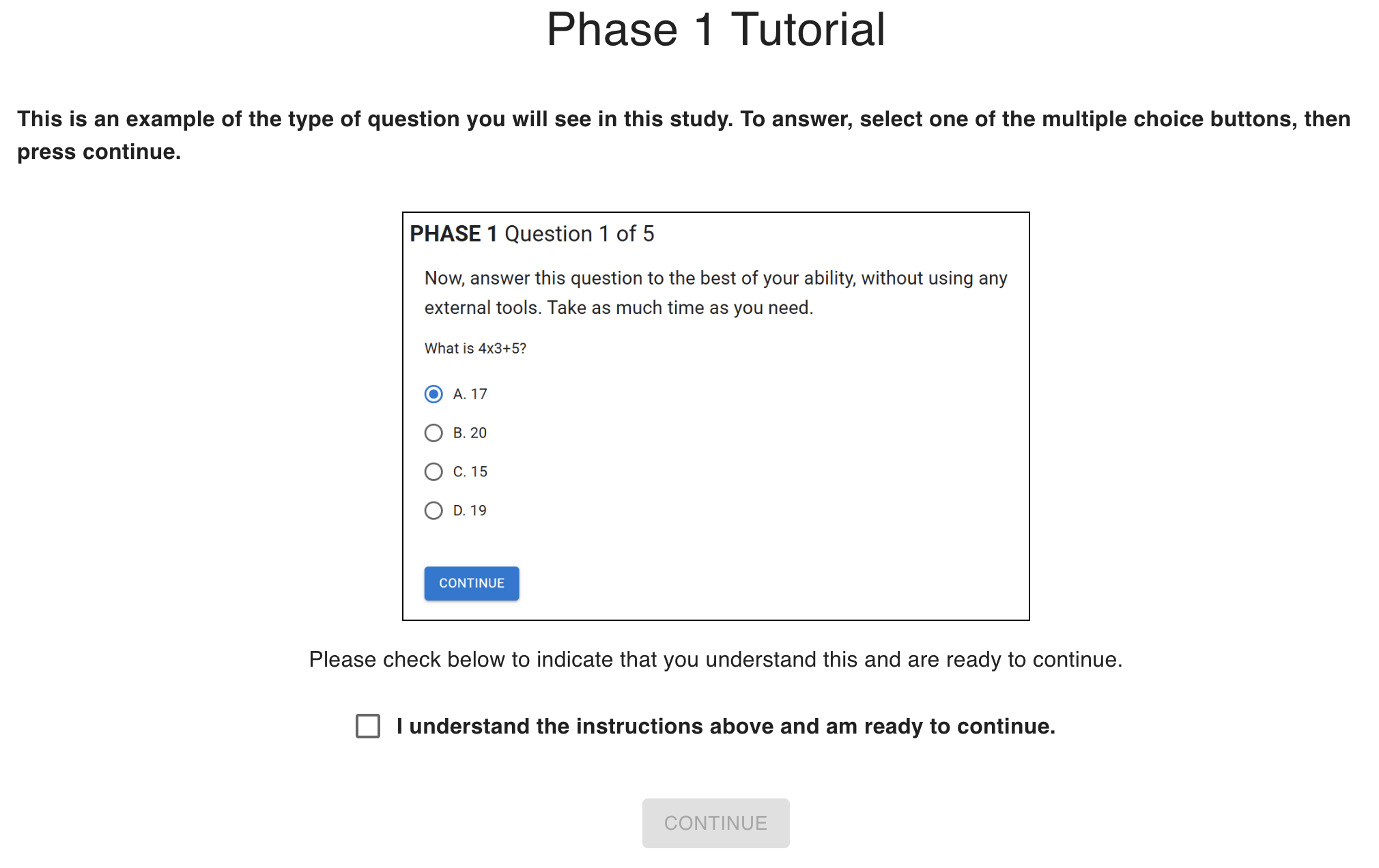}
    \caption{Phase 1 Tutorial. Provides an example of a Phase 1 question before the user begins Phase 1.}
    \label{fig:phase1_tutorial}
\end{figure*}

\begin{figure*}
    \centering
    \includegraphics[width=0.7\linewidth]{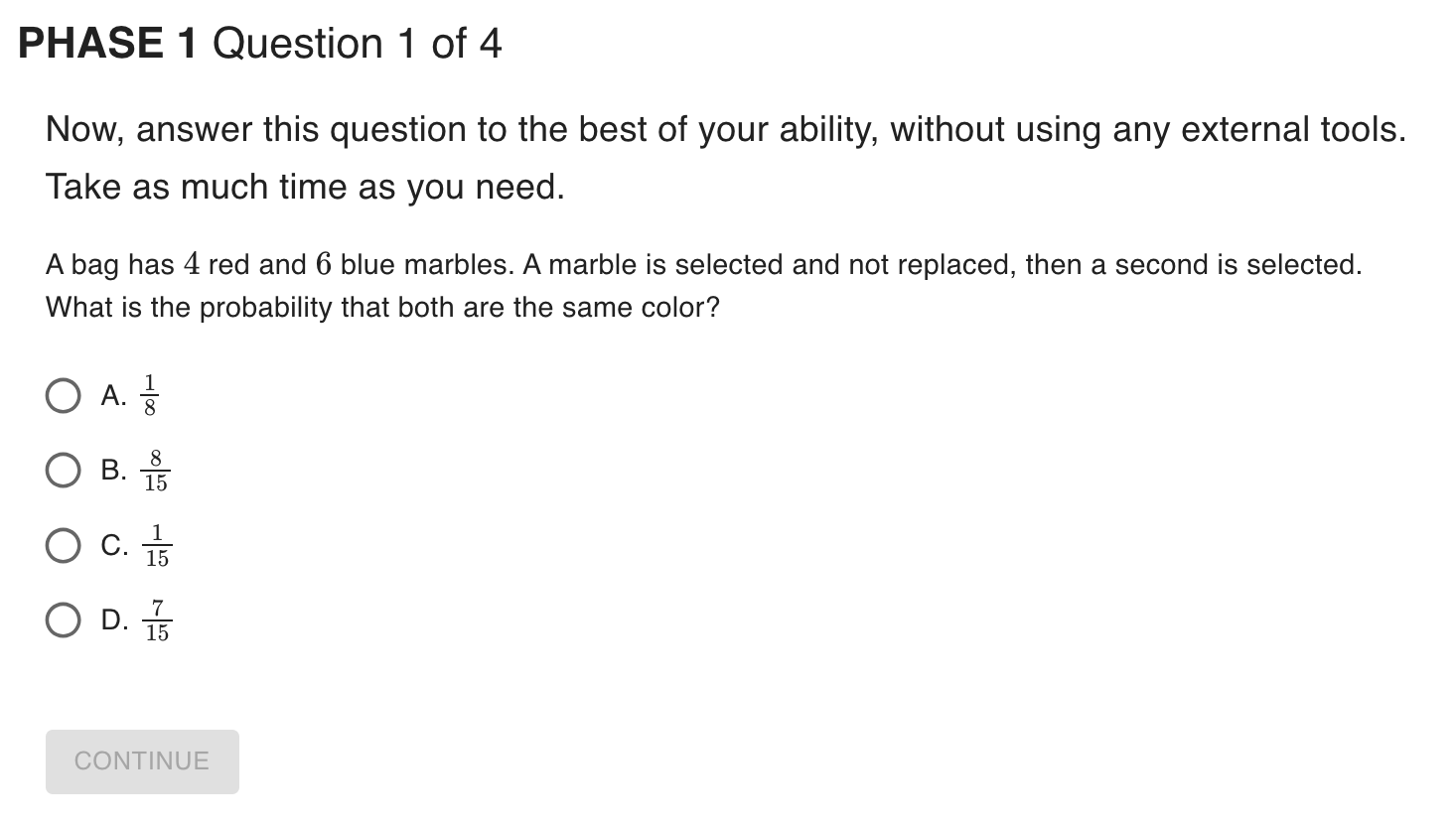}
    \caption{User-alone page. Users answer all questions on their own in Phase 1 and, if they are in the \textit{answer-first} condition, answer each question in Phase 2 on their own first before answering with AI.}
    \label{fig:user_alone}
\end{figure*}

\begin{figure*}
    \centering
    \includegraphics[width=0.9\linewidth]{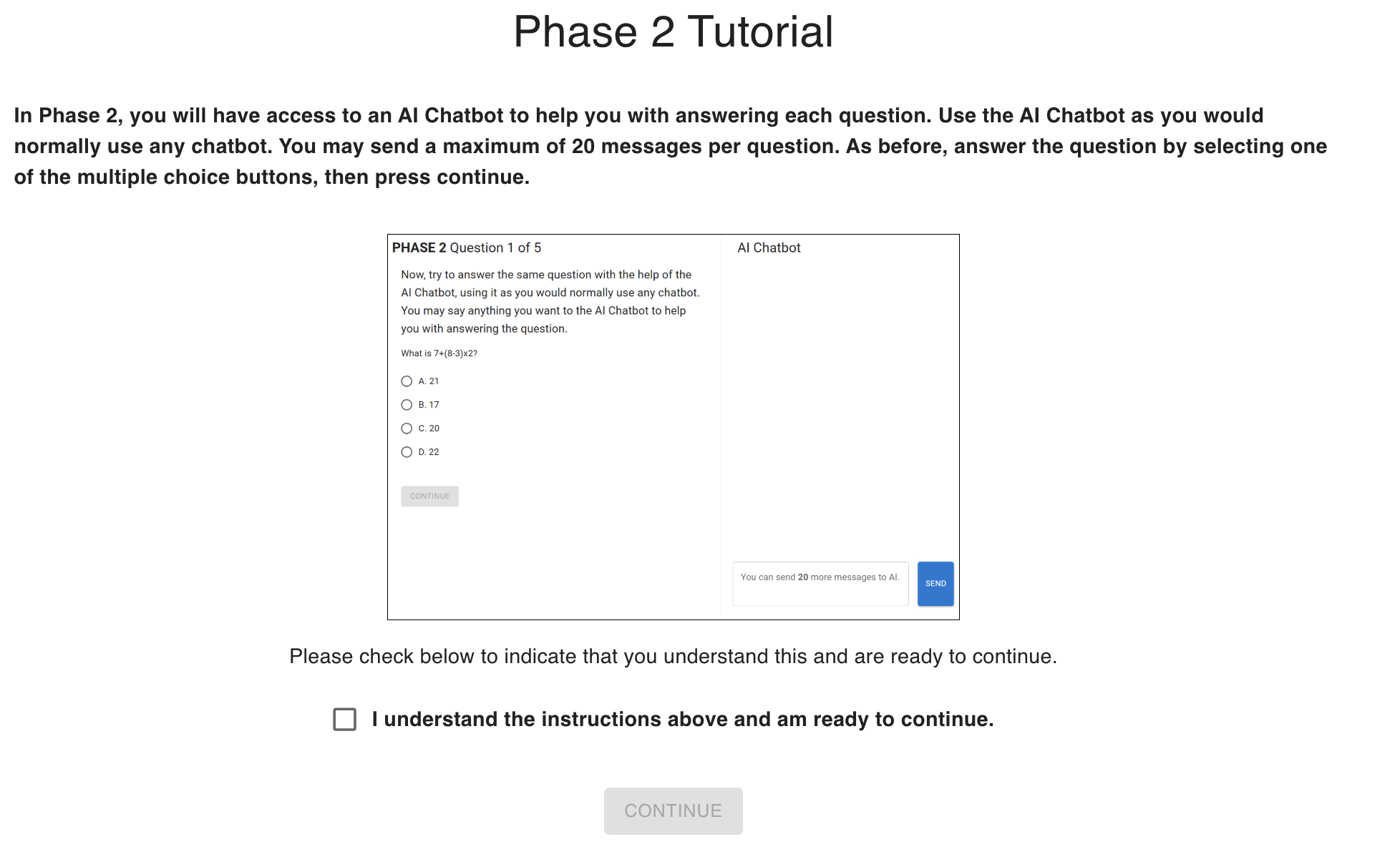}
    \caption{Phase 2 Tutorial. Provides an example of a Phase 2 question before the user begins Phase 2.}
    \label{fig:phase2_tutorial}
\end{figure*}

\begin{figure*}
    \centering
    \includegraphics[width=0.8\linewidth]{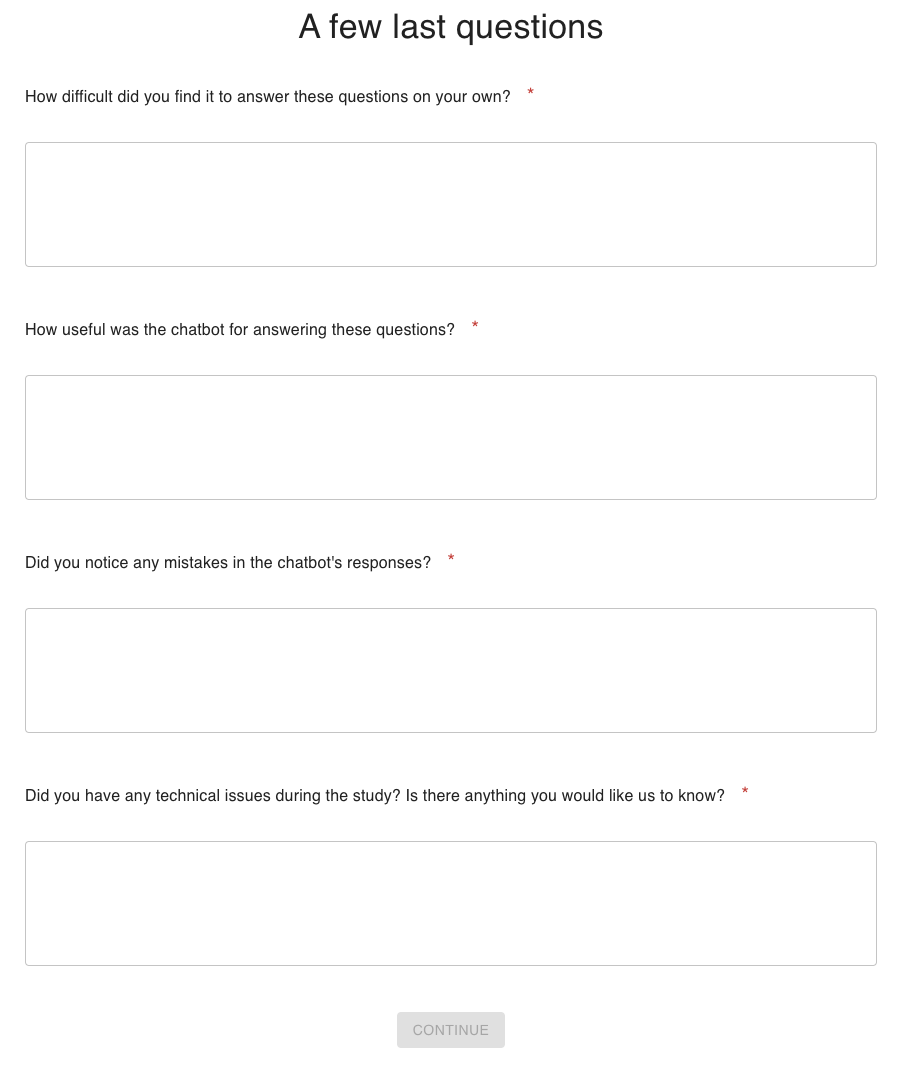}
    \caption{Feedback page. Final page of the task, where users leave free-text feedback to various questions.}
    \label{fig:feedback}
\end{figure*}

\paragraph{Interface.}
We implemented our app in React and used Azure CosmosDB as our database. 
We provide screenshots of all of the pages in our user study interface, including the Introduction Page (Figure \ref{fig:introduction}), Phase 1 Tutorial (Figure \ref{fig:phase1_tutorial}), Confidence Page (Figure \ref{fig:confidence}), User-Alone Page (Figure \ref{fig:user_alone}), Phase 2 Instructions (Figure \ref{fig:phase2_instructions}), Phase 2 Tutorial (Figure \ref{fig:phase2_tutorial}), User-AI Page (Figure \ref{fig:phase2_screenshot}), and Feedback Page (Figure \ref{fig:feedback}).

On the User-AI Page (Figure \ref{fig:phase2_screenshot}), we tried to minimize our influence on the user-AI interactions, but we also wanted to ensure that the users put in meaningful effort to interact with AI, as they would if they were intrinsically motivated to answer a question.
We required the user to send at least one message before moving onto the next question and we said in the instructions that low-effort conversations, e.g., only ``hi'', would be flagged. 
We also disabled copy-and-paste of the question text, both to prevent the use of external tools (e.g., ChatGPT) and to prevent trivial conversations where the user simply copy-and-pasted. 
While users may copy-and-paste in the real world if presented with a question, we were trying to capture the case where a user had intrinsic motivation to answer a question, and in those cases, the question would usually be in their heads so they would not have something to copy-and-paste. 
Furthermore, our \textit{free-text} AI-alone method already serves as a good estimate of user copy-and-paste, since it simply copy-and-pastes the question to the AI as the first prompt, then uses GPT-4o to extract an answer from the AI's free-text response. 

\include{tables/mean_accs_gpt4o}
\include{tables/mean_accs_llama8b}

\subsection{Pilots and Incentivization}
\label{sec:incentivization}

\paragraph{Pilot 1: no incentives.}
We ran one medium-sized pilot with 100 participants where we tested all datasets and models.
At this point, we also included GPT-4o-mini as a third model, in addition to GPT-4o and Llama-3.1-8b.
In this pilot, we did not include incentives for correct answers.
Results from this pilot did not show significant differences in accuracy between GPT-4o and GPT-4o-mini, so we decided to drop GPT-4o-mini from our full study, so that we could gather more answers per model.

\paragraph{Pilot 2: testing incentives.}
In our second pilot, we wanted to test the effect of including a small incentive for getting the correct answer, hypothesizing that it might improve the ecological validity of the study since users would try harder to answer the questions correctly.
We included a small bonus of \$0.10 per correct answer, with a maximum bonus of \$1.30 for 13 questions, on top of the same base compensation of \$5.00 for completing the task.

\begin{figure}
    \centering
    \includegraphics[width=\linewidth]{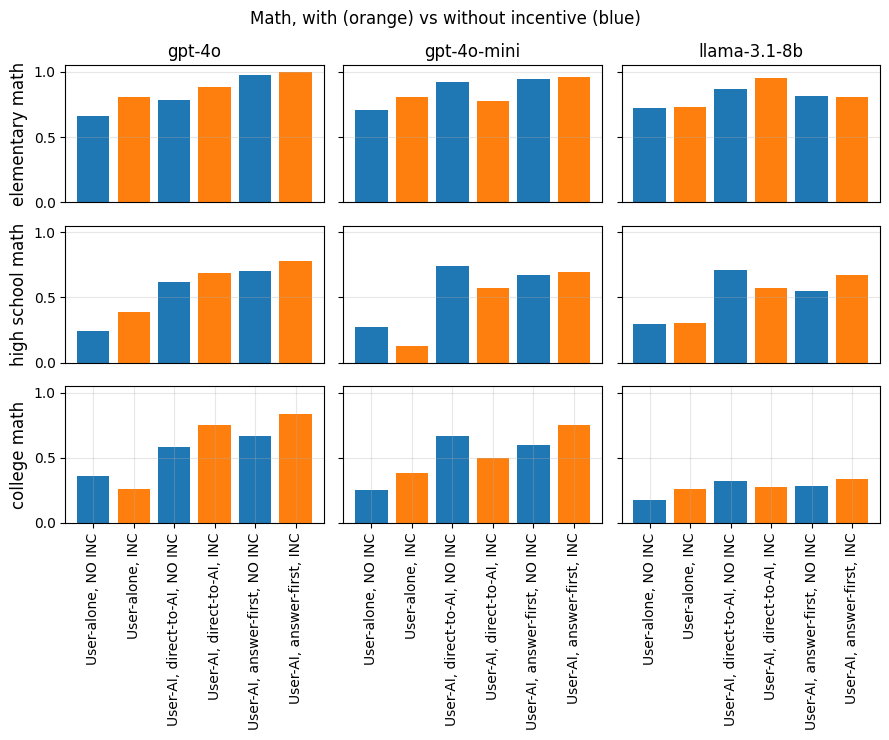}
    \caption{Comparing results from Pilot 1 (without incentives) and Pilot 2 (with incentives).}
    \label{fig:pilot_comparison}
\end{figure}
\begin{figure*}
    \centering
    \includegraphics[width=0.9\linewidth]{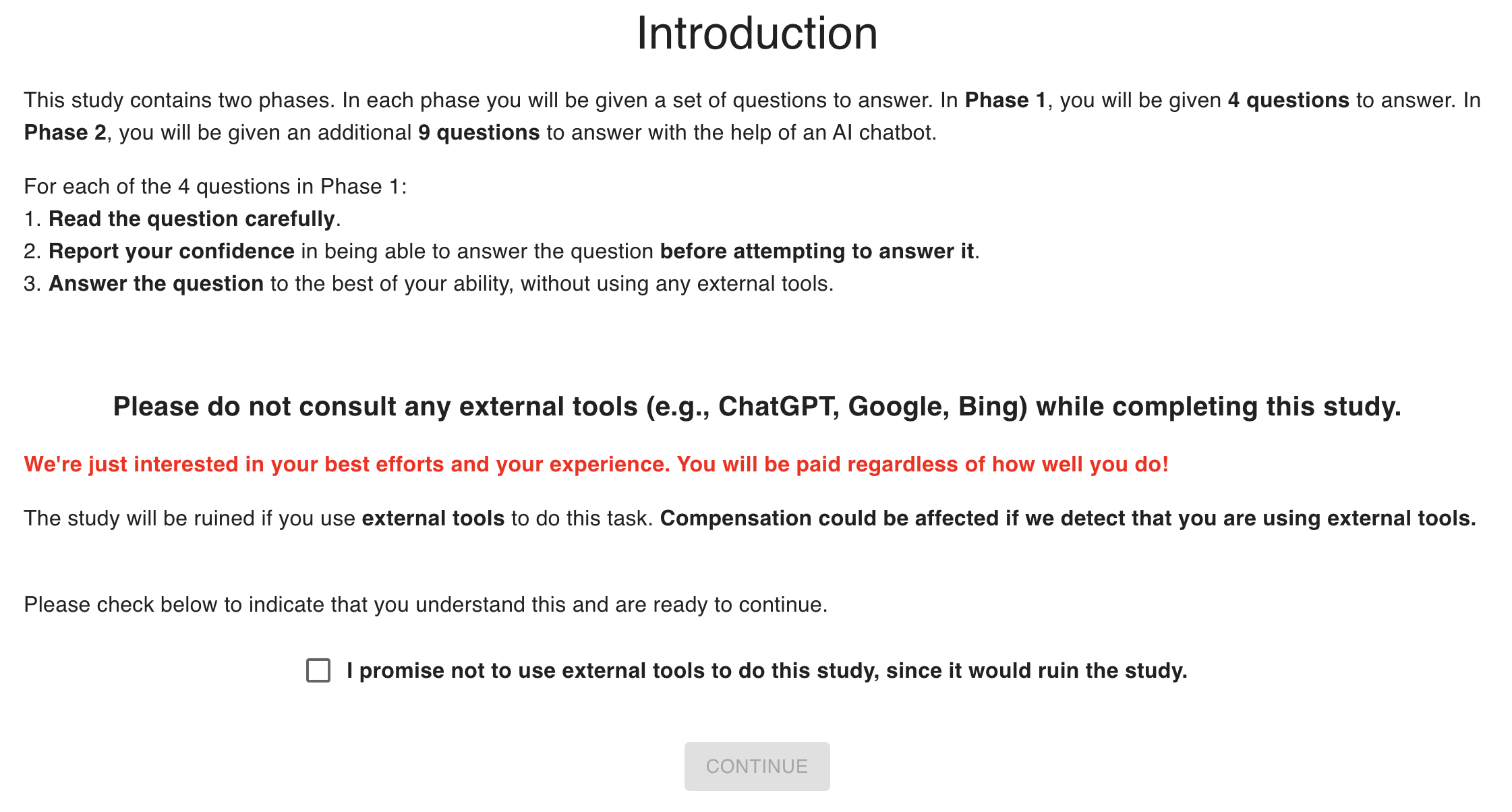}
    \caption{Introduction page. Explains the task to users and ensures that they do not consult external tools.}
    \label{fig:introduction}
\end{figure*}
\begin{figure*}
    \centering
    \includegraphics[width=0.9\linewidth]{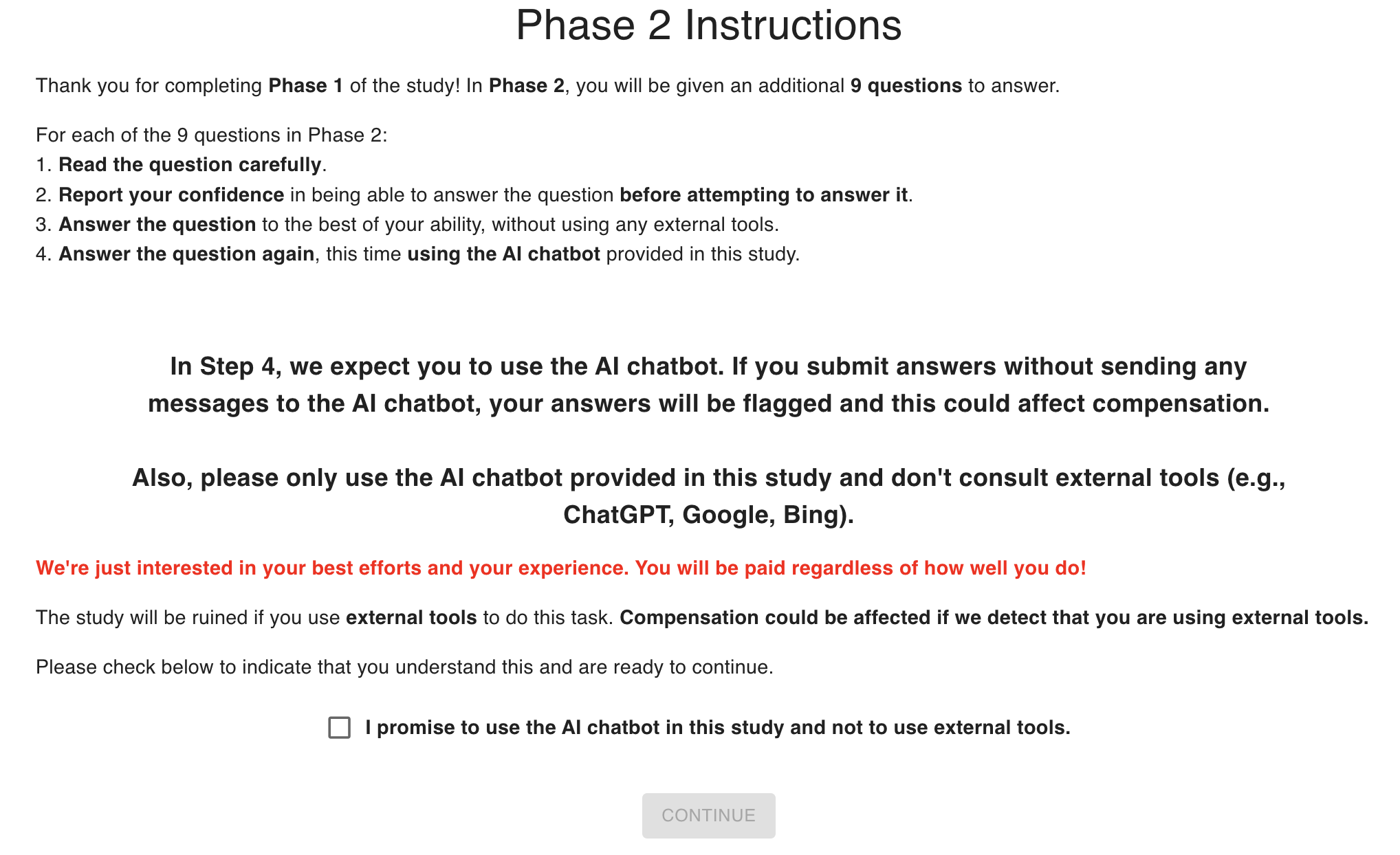}
    \caption{Phase 2 Instructions. Explains to users what they can expect in Phase 2 and reminds them not to use external tools. This screenshot shows instructions for a user in the \textit{answer-first} condition. Users in the \textit{direct-to-AI} condition see similar instructions, but without Step 3.}
    \label{fig:phase2_instructions}
\end{figure*}

While this bonus could help to improve ecological validity, there was a risk that the incentives result in users cheating on the study, such as by searching for the question on Google or ChatGPT.
To mitigate this risk, first we repeatedly required users to acknowledge that they would not use external tools (Figures \ref{fig:introduction} and \ref{fig:phase2_instructions}) and we said, ``Compensation could be affected if we detect that you are using external tool.''
Second, we ran a second medium-sized pilot with incentives, with 60 participants on the three math datasets, and we compared the results between Pilots 1 and 2 to see if Pilot 2 had unrealistic increases in accuracy that could not be explained by slightly more user effort.

We visualize the mean accuracies per dataset and model in Figure \ref{fig:pilot_comparison}. 
We found that, as expected, incentives tended to improve performance a little: out of 27 combinations of math datasets (3), models (3), and answer types (i.e., user-alone, user-AI \textit{answer-first}, and user-AI \textit{direct-to-AI}), the pilot with incentives had a higher mean accuracy 19 times.
We also found that conversations were slightly longer with incentives.
However, the overall improvement in accuracy was very small, only 3 percentage points, meaning we did not see unrealistic improvements that would suggest use of external tools.
We also continued to see the gaps in user-AI performance between the GPT models and Llama-3.1-8b, suggesting users were basing their answers on the AI Chatbot given to them.
As further evidence of the use of the AI Chatbot, and not external tools, we found that in the vast majority of cases (63 out of 66 examples) where the user \textit{changed} from an incorrect user-alone answer to a correct user-AI answer, that new answer matched the answer given by the AI model in the user-AI conversation.
Since we found that incentives seemed to encourage users to try slightly harder, and we did not see evidence of cheating, we decided to keep incentives for our full study, but our pilot comparison shows that our results were not overly sensitive to this decision.

\subsection{\dataname{}}
\begin{figure*}
    \centering
    \includegraphics[width=0.7\linewidth]{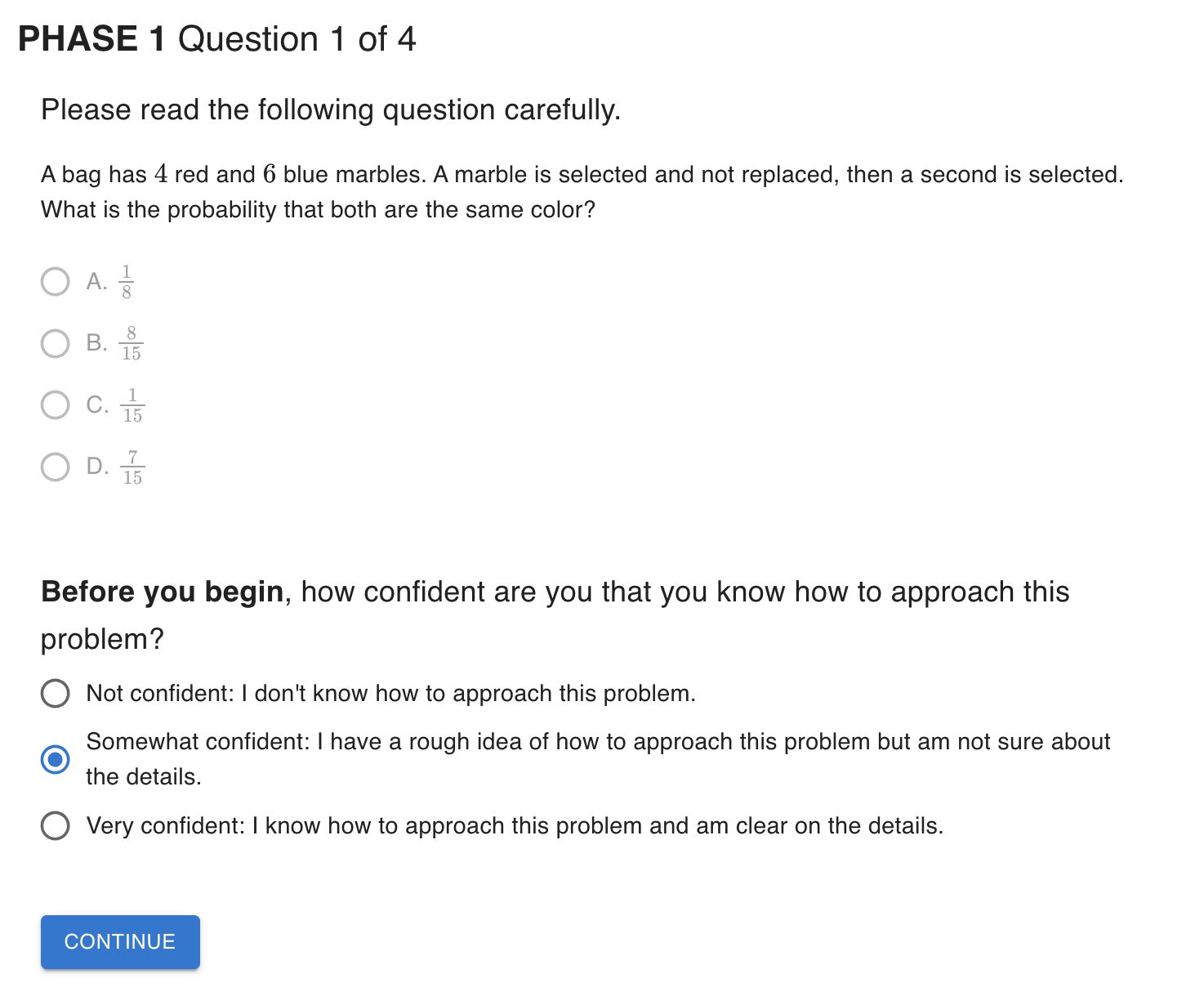}
    \caption{Confidence page. Included per-question in both phases before the user tries to answer each question.}
    \label{fig:confidence}
\end{figure*}

Our dataset, \dataname{}\footnote{\huggingface{}.}, compiles data over the full study (650 participants, with incentives) and the two pilots.
\dataname{} consists of user-alone answers, user-AI answers and conversations, and user confidence, which we asked users to report per question before the attempting the question (Figure~\ref{fig:confidence}).
\dataname{} also includes AI-alone answers from our AI-alone experiments, where we tested each model 50 times per question and AI-alone method (see Section~\ref{sec:ai_alone_experiments} for details).

In total, \dataname{} contains 7,148 user-alone answers, \numConversations{} user-AI answers and conversations, 10,828 user confidence answers, and 118,717 AI-alone answers, resulting in 144,029 answers in total.
\dataname{} contains data from more than the total number of participants we recruited (810 = 650 + 100 + 60), since some participants started but did not complete the study.
In Table \ref{tab:data_stats}, we provide additional data statistics, including how many answers we collected per model, dataset, condition, and answer type (user-alone or user-AI).

\paragraph{Protecting participant privacy.}
To protect the privacy of our participants, first we mapped each person's Prolific ID to a new, randomly generated string of 10 letters and digits, checking that there were no collisions between individuals. 
The \texttt{worker\_id} field in \dataname{} contains these new strings, instead of their original worker ID on Prolific.
Second, we checked for personally identifying information in the user-AI conversations.
We used the EU GDPR's definition of personal data:\footnote{\url{https://gdpr.eu/eu-gdpr-personal-data/}.}
\begin{quote}
‘Personal data’ means any information relating to an identified or identifiable natural person (‘data subject’); an identifiable natural person is one who can be identified, directly or indirectly, in particular by reference to an identifier such as a name, an identification number, location data, an online identifier or to one or more factors specific to the physical, physiological, genetic, mental, economic, cultural or social identity of that natural person.
\end{quote}
To check if a participant had revealed personal information, we provided our private instance of GPT-4o with all of their user-AI conversations, along with GDPR's definition of personal data, and prompted GPT-4o to answer whether the conversations contained any personal data.

As expected, given the nature of these conversations (answering benchmark questions), there were very few conversations with personal data.
Over 823 participants with user-AI conversations, GPT-4o only flagged three participants with personal data.
We manually inspected these three and found that two were not actually revealing personal data; they were both rephrasing a math question, ``Carlos Montado was born on Saturday, November 9, 2002'', in first person, leading GPT-4o to think that they were providing their birthday. 
One participant appeared to share personal details, so we removed their conversations from our public release.

\paragraph{Filtering from \dataname{} for statistical analysis.}
For our main statistical analyses (Section~\ref{sec:study-results}), we only used data from the full study, and not from the two pilots.
We furthermore filtered the data following the criteria we described in our preregistration,\footnote{\url{https://aspredicted.org/n84n-sn3f.pdf.}} such as only keeping the workers' answers from their first assignment if they had multiple.
While we could control from Prolific that workers could not participate in our task multiple times, once they opened our app, they could start the study then be taken back to the beginning of the study flow (Figure~\ref{fig:flow}) if they refreshed the app.
If they did so, we would not want to keep their data after they refreshed, since their behavior on the second time around could be affected by what they already saw the first time.
When a worker opened the app or refreshed, they received a new assignment, defined by their combination of model (one of 2), user-AI condition (one of 2), subject (one of 3), and question batch (one of 7 or 18). 
The probability that they would receive the same assignment twice if they refreshed was very low (less than 1\%), so we could check for multiple assignments to test whether they refreshed, and we used timestamps to determine their first assignment.
Ultimately, we found very few workers (3\%) with multiple assignments, and for those workers, we kept their data from their first assignment. 

Overall, our filtering criteria was as follows:
\begin{itemize}
\item First, we only kept data from the workers who completed the study, which we checked by cross-referencing the list of worker IDs given to us by Prolific of all workers who completed the study and clicked on the completion code.
\item Second, we only kept data from workers who passed the attention check. The vast majority of workers (99.5\%) passed the attention check. We also did not include answers from the attention check in our user-alone estimates, since the attention check was easier than the other questions.
\item Third, we checked for workers with multiple assignments, and we only kept data from those workers on their first assignment.
\end{itemize}
In our released version of \dataname{}, we include all the data from the full study without filtering and the filtered version of the data, for replicability and transparency.

\include{tables/chatbench_stats}

\begin{figure*}
    \centering
    \includegraphics[width=0.9\linewidth]{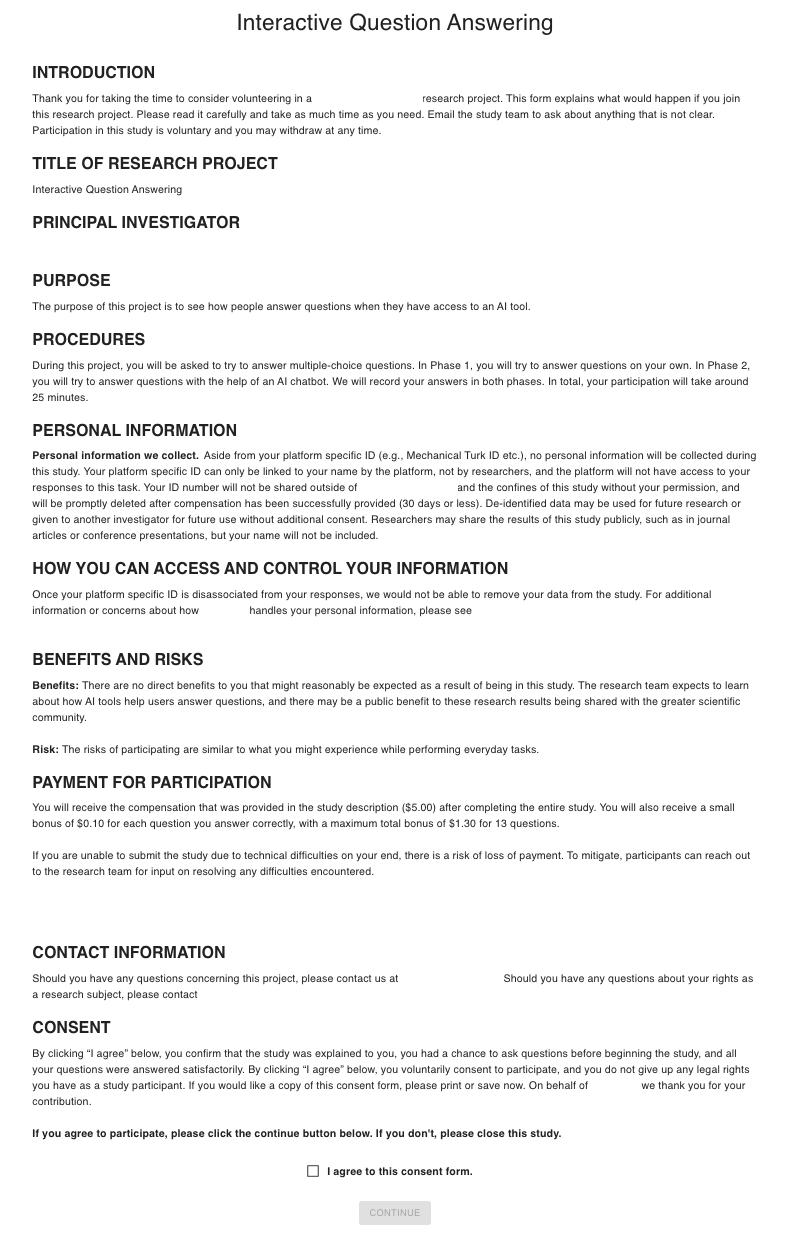}
    \caption{Consent page. Parts are redacted to remain anonymous.}
    \label{fig:consent}
\end{figure*}

\clearpage

%% file: tables/mean_accs_gpt4o.tex
\begin{table*}[t]
\begin{adjustwidth}{-1cm}{-1cm}
\centering
\small
\begin{tabular}[t]{lllrrrrrr}
\toprule
Dataset & Model & Comparison & $\mathrm{Acc}_1$ & $\mathrm{SE}_1$ & $\mathrm{Acc}_2$ & $\mathrm{SE}_2$ & $z$-value & $p$-value\\
\midrule
\cellcolor{gray!10}{Elementary Math} & \cellcolor{gray!10}{GPT-4o} & \cellcolor{gray!10}{AI letter zero shot vs. UserAI direct to ai} & \cellcolor{gray!10}{0.73} & \cellcolor{gray!10}{0.04} & \cellcolor{gray!10}{0.92} & \cellcolor{gray!10}{0.02} & \cellcolor{gray!10}{-3.92} & \cellcolor{gray!10}{$<$\textbf{0.01}}\\
Elementary Math & GPT-4o & AI letter zero shot vs. UserAI answer first & 0.73 & 0.04 & 0.90 & 0.02 & -3.43 & $<$\textbf{0.01}\\
\cellcolor{gray!10}{Elementary Math} & \cellcolor{gray!10}{GPT-4o} & \cellcolor{gray!10}{AI letter few shot vs. UserAI direct to ai} & \cellcolor{gray!10}{0.74} & \cellcolor{gray!10}{0.04} & \cellcolor{gray!10}{0.92} & \cellcolor{gray!10}{0.02} & \cellcolor{gray!10}{-3.83} & \cellcolor{gray!10}{$<$\textbf{0.01}}\\
Elementary Math & GPT-4o & AI letter few shot vs. UserAI answer first & 0.74 & 0.04 & 0.90 & 0.02 & -3.34 & $<$\textbf{0.01}\\
\cellcolor{gray!10}{Elementary Math} & \cellcolor{gray!10}{GPT-4o} & \cellcolor{gray!10}{AI free text vs. UserAI direct to ai} & \cellcolor{gray!10}{0.99} & \cellcolor{gray!10}{0.01} & \cellcolor{gray!10}{0.92} & \cellcolor{gray!10}{0.02} & \cellcolor{gray!10}{3.03} & \cellcolor{gray!10}{$<$\textbf{0.01}}\\
\addlinespace
Elementary Math & GPT-4o & AI free text vs. UserAI answer first & 0.99 & 0.01 & 0.90 & 0.02 & 4.04 & $<$\textbf{0.01}\\
\cellcolor{gray!10}{Elementary Math} & \cellcolor{gray!10}{GPT-4o} & \cellcolor{gray!10}{User alone vs. UserAI direct to ai} & \cellcolor{gray!10}{0.78} & \cellcolor{gray!10}{0.03} & \cellcolor{gray!10}{0.92} & \cellcolor{gray!10}{0.02} & \cellcolor{gray!10}{-4.21} & \cellcolor{gray!10}{$<$\textbf{0.01}}\\
Elementary Math & GPT-4o & User alone vs. UserAI answer first & 0.78 & 0.03 & 0.90 & 0.02 & -3.52 & $<$\textbf{0.01}\\
\cellcolor{gray!10}{High School Math} & \cellcolor{gray!10}{GPT-4o} & \cellcolor{gray!10}{AI letter zero shot vs. UserAI direct to ai} & \cellcolor{gray!10}{0.51} & \cellcolor{gray!10}{0.05} & \cellcolor{gray!10}{0.70} & \cellcolor{gray!10}{0.04} & \cellcolor{gray!10}{-3.20} & \cellcolor{gray!10}{$<$\textbf{0.01}}\\
High School Math & GPT-4o & AI letter zero shot vs. UserAI answer first & 0.51 & 0.05 & 0.73 & 0.03 & -3.92 & $<$\textbf{0.01}\\
\addlinespace
\cellcolor{gray!10}{High School Math} & \cellcolor{gray!10}{GPT-4o} & \cellcolor{gray!10}{AI letter few shot vs. UserAI direct to ai} & \cellcolor{gray!10}{0.49} & \cellcolor{gray!10}{0.04} & \cellcolor{gray!10}{0.70} & \cellcolor{gray!10}{0.04} & \cellcolor{gray!10}{-3.57} & \cellcolor{gray!10}{$<$\textbf{0.01}}\\
High School Math & GPT-4o & AI letter few shot vs. UserAI answer first & 0.49 & 0.04 & 0.73 & 0.03 & -4.33 & $<$\textbf{0.01}\\
\cellcolor{gray!10}{High School Math} & \cellcolor{gray!10}{GPT-4o} & \cellcolor{gray!10}{AI free text vs. UserAI direct to ai} & \cellcolor{gray!10}{0.85} & \cellcolor{gray!10}{0.03} & \cellcolor{gray!10}{0.70} & \cellcolor{gray!10}{0.04} & \cellcolor{gray!10}{3.14} & \cellcolor{gray!10}{$<$\textbf{0.01}}\\
High School Math & GPT-4o & AI free text vs. UserAI answer first & 0.85 & 0.03 & 0.73 & 0.03 & 2.73 & $<$\textbf{0.01}\\
\cellcolor{gray!10}{High School Math} & \cellcolor{gray!10}{GPT-4o} & \cellcolor{gray!10}{User alone vs. UserAI direct to ai} & \cellcolor{gray!10}{0.41} & \cellcolor{gray!10}{0.03} & \cellcolor{gray!10}{0.70} & \cellcolor{gray!10}{0.04} & \cellcolor{gray!10}{-5.88} & \cellcolor{gray!10}{$<$\textbf{0.01}}\\
\addlinespace
High School Math & GPT-4o & User alone vs. UserAI answer first & 0.41 & 0.03 & 0.73 & 0.03 & -7.03 & $<$\textbf{0.01}\\
\cellcolor{gray!10}{College Math} & \cellcolor{gray!10}{GPT-4o} & \cellcolor{gray!10}{AI letter zero shot vs. UserAI direct to ai} & \cellcolor{gray!10}{0.45} & \cellcolor{gray!10}{0.07} & \cellcolor{gray!10}{0.52} & \cellcolor{gray!10}{0.08} & \cellcolor{gray!10}{-0.61} & \cellcolor{gray!10}{0.54}\\
College Math & GPT-4o & AI letter zero shot vs. UserAI answer first & 0.45 & 0.07 & 0.52 & 0.07 & -0.72 & 0.47\\
\cellcolor{gray!10}{College Math} & \cellcolor{gray!10}{GPT-4o} & \cellcolor{gray!10}{AI letter few shot vs. UserAI direct to ai} & \cellcolor{gray!10}{0.44} & \cellcolor{gray!10}{0.07} & \cellcolor{gray!10}{0.52} & \cellcolor{gray!10}{0.08} & \cellcolor{gray!10}{-0.72} & \cellcolor{gray!10}{0.47}\\
College Math & GPT-4o & AI letter few shot vs. UserAI answer first & 0.44 & 0.07 & 0.52 & 0.07 & -0.85 & 0.40\\
\addlinespace
\cellcolor{gray!10}{College Math} & \cellcolor{gray!10}{GPT-4o} & \cellcolor{gray!10}{AI free text vs. UserAI direct to ai} & \cellcolor{gray!10}{0.73} & \cellcolor{gray!10}{0.06} & \cellcolor{gray!10}{0.52} & \cellcolor{gray!10}{0.08} & \cellcolor{gray!10}{2.23} & \cellcolor{gray!10}{\textbf{0.03}}\\
College Math & GPT-4o & AI free text vs. UserAI answer first & 0.73 & 0.06 & 0.52 & 0.07 & 2.40 & \textbf{0.02}\\
\cellcolor{gray!10}{College Math} & \cellcolor{gray!10}{GPT-4o} & \cellcolor{gray!10}{User alone vs. UserAI direct to ai} & \cellcolor{gray!10}{0.28} & \cellcolor{gray!10}{0.04} & \cellcolor{gray!10}{0.52} & \cellcolor{gray!10}{0.08} & \cellcolor{gray!10}{-2.67} & \cellcolor{gray!10}{$<$\textbf{0.01}}\\
College Math & GPT-4o & User alone vs. UserAI answer first & 0.28 & 0.04 & 0.52 & 0.07 & -3.10 & $<$\textbf{0.01}\\
\cellcolor{gray!10}{Conceptual Physics} & \cellcolor{gray!10}{GPT-4o} & \cellcolor{gray!10}{AI letter zero shot vs. UserAI direct to ai} & \cellcolor{gray!10}{0.91} & \cellcolor{gray!10}{0.03} & \cellcolor{gray!10}{0.84} & \cellcolor{gray!10}{0.03} & \cellcolor{gray!10}{1.74} & \cellcolor{gray!10}{0.08}\\
\addlinespace
Conceptual Physics & GPT-4o & AI letter zero shot vs. UserAI answer first & 0.91 & 0.03 & 0.84 & 0.03 & 1.70 & 0.09\\
\cellcolor{gray!10}{Conceptual Physics} & \cellcolor{gray!10}{GPT-4o} & \cellcolor{gray!10}{AI letter few shot vs. UserAI direct to ai} & \cellcolor{gray!10}{0.96} & \cellcolor{gray!10}{0.02} & \cellcolor{gray!10}{0.84} & \cellcolor{gray!10}{0.03} & \cellcolor{gray!10}{3.22} & \cellcolor{gray!10}{$<$\textbf{0.01}}\\
Conceptual Physics & GPT-4o & AI letter few shot vs. UserAI answer first & 0.96 & 0.02 & 0.84 & 0.03 & 3.22 & $<$\textbf{0.01}\\
\cellcolor{gray!10}{Conceptual Physics} & \cellcolor{gray!10}{GPT-4o} & \cellcolor{gray!10}{AI free text vs. UserAI direct to ai} & \cellcolor{gray!10}{0.97} & \cellcolor{gray!10}{0.02} & \cellcolor{gray!10}{0.84} & \cellcolor{gray!10}{0.03} & \cellcolor{gray!10}{3.62} & \cellcolor{gray!10}{$<$\textbf{0.01}}\\
Conceptual Physics & GPT-4o & AI free text vs. UserAI answer first & 0.97 & 0.02 & 0.84 & 0.03 & 3.63 & $<$\textbf{0.01}\\
\addlinespace
\cellcolor{gray!10}{Conceptual Physics} & \cellcolor{gray!10}{GPT-4o} & \cellcolor{gray!10}{User alone vs. UserAI direct to ai} & \cellcolor{gray!10}{0.55} & \cellcolor{gray!10}{0.03} & \cellcolor{gray!10}{0.84} & \cellcolor{gray!10}{0.03} & \cellcolor{gray!10}{-6.48} & \cellcolor{gray!10}{$<$\textbf{0.01}}\\
Conceptual Physics & GPT-4o & User alone vs. UserAI answer first & 0.55 & 0.03 & 0.84 & 0.03 & -6.69 & $<$\textbf{0.01}\\
\cellcolor{gray!10}{Moral Scenarios} & \cellcolor{gray!10}{GPT-4o} & \cellcolor{gray!10}{AI letter zero shot vs. UserAI direct to ai} & \cellcolor{gray!10}{0.71} & \cellcolor{gray!10}{0.05} & \cellcolor{gray!10}{0.79} & \cellcolor{gray!10}{0.03} & \cellcolor{gray!10}{-1.47} & \cellcolor{gray!10}{0.14}\\
Moral Scenarios & GPT-4o & AI letter zero shot vs. UserAI answer first & 0.71 & 0.05 & 0.78 & 0.04 & -1.13 & 0.26\\
\cellcolor{gray!10}{Moral Scenarios} & \cellcolor{gray!10}{GPT-4o} & \cellcolor{gray!10}{AI letter few shot vs. UserAI direct to ai} & \cellcolor{gray!10}{0.80} & \cellcolor{gray!10}{0.04} & \cellcolor{gray!10}{0.79} & \cellcolor{gray!10}{0.03} & \cellcolor{gray!10}{0.27} & \cellcolor{gray!10}{0.79}\\
\addlinespace
Moral Scenarios & GPT-4o & AI letter few shot vs. UserAI answer first & 0.80 & 0.04 & 0.78 & 0.04 & 0.49 & 0.63\\
\cellcolor{gray!10}{Moral Scenarios} & \cellcolor{gray!10}{GPT-4o} & \cellcolor{gray!10}{AI free text vs. UserAI direct to ai} & \cellcolor{gray!10}{0.72} & \cellcolor{gray!10}{0.05} & \cellcolor{gray!10}{0.79} & \cellcolor{gray!10}{0.03} & \cellcolor{gray!10}{-1.26} & \cellcolor{gray!10}{0.21}\\
Moral Scenarios & GPT-4o & AI free text vs. UserAI answer first & 0.72 & 0.05 & 0.78 & 0.04 & -0.93 & 0.35\\
\cellcolor{gray!10}{Moral Scenarios} & \cellcolor{gray!10}{GPT-4o} & \cellcolor{gray!10}{User alone vs. UserAI direct to ai} & \cellcolor{gray!10}{0.73} & \cellcolor{gray!10}{0.03} & \cellcolor{gray!10}{0.79} & \cellcolor{gray!10}{0.03} & \cellcolor{gray!10}{-1.54} & \cellcolor{gray!10}{0.12}\\
Moral Scenarios & GPT-4o & User alone vs. UserAI answer first & 0.73 & 0.03 & 0.78 & 0.04 & -1.05 & 0.29\\
\bottomrule
\end{tabular}
\caption{Results per dataset for GPT-4o, including AI-alone vs. user-AI comparisons and user-alone vs. user-AI comparisons.}
\label{tab:gpt4o-results}
\end{adjustwidth}
\end{table*}

%% file: tables/mean_accs_llama8b.tex
\begin{table*}[t]
\begin{adjustwidth}{-1cm}{-1cm}
\centering
\small
\begin{tabular}[t]{lllrrrrrr}
\toprule
Dataset & Model & Comparison & $\mathrm{Acc}_1$ & $\mathrm{SE}_1$ & $\mathrm{Acc}_2$ & $\mathrm{SE}_2$ & $z$-value & $p$-value\\
\midrule
\cellcolor{gray!10}{Elementary Math} & \cellcolor{gray!10}{Llama-3.1-8b} & \cellcolor{gray!10}{AI letter zero shot vs. UserAI direct to ai} & \cellcolor{gray!10}{0.45} & \cellcolor{gray!10}{0.04} & \cellcolor{gray!10}{0.86} & \cellcolor{gray!10}{0.03} & \cellcolor{gray!10}{-8.58} & \cellcolor{gray!10}{$<$\textbf{0.01}}\\
Elementary Math & Llama-3.1-8b & AI letter zero shot vs. UserAI answer first & 0.45 & 0.04 & 0.90 & 0.02 & -10.50 & $<$\textbf{0.01}\\
\cellcolor{gray!10}{Elementary Math} & \cellcolor{gray!10}{Llama-3.1-8b} & \cellcolor{gray!10}{AI letter few shot vs. UserAI direct to ai} & \cellcolor{gray!10}{0.43} & \cellcolor{gray!10}{0.03} & \cellcolor{gray!10}{0.86} & \cellcolor{gray!10}{0.03} & \cellcolor{gray!10}{-9.39} & \cellcolor{gray!10}{$<$\textbf{0.01}}\\
Elementary Math & Llama-3.1-8b & AI letter few shot vs. UserAI answer first & 0.43 & 0.03 & 0.90 & 0.02 & -11.53 & $<$\textbf{0.01}\\
\cellcolor{gray!10}{Elementary Math} & \cellcolor{gray!10}{Llama-3.1-8b} & \cellcolor{gray!10}{AI free text vs. UserAI direct to ai} & \cellcolor{gray!10}{0.88} & \cellcolor{gray!10}{0.03} & \cellcolor{gray!10}{0.86} & \cellcolor{gray!10}{0.03} & \cellcolor{gray!10}{0.56} & \cellcolor{gray!10}{0.58}\\
\addlinespace
Elementary Math & Llama-3.1-8b & AI free text vs. UserAI answer first & 0.88 & 0.03 & 0.90 & 0.02 & -0.65 & 0.51\\
\cellcolor{gray!10}{Elementary Math} & \cellcolor{gray!10}{Llama-3.1-8b} & \cellcolor{gray!10}{User alone vs. UserAI direct to ai} & \cellcolor{gray!10}{0.81} & \cellcolor{gray!10}{0.03} & \cellcolor{gray!10}{0.86} & \cellcolor{gray!10}{0.03} & \cellcolor{gray!10}{-1.26} & \cellcolor{gray!10}{0.21}\\
Elementary Math & Llama-3.1-8b & User alone vs. UserAI answer first & 0.81 & 0.03 & 0.90 & 0.02 & -2.70 & $<$\textbf{0.01}\\
\cellcolor{gray!10}{High School Math} & \cellcolor{gray!10}{Llama-3.1-8b} & \cellcolor{gray!10}{AI letter zero shot vs. UserAI direct to ai} & \cellcolor{gray!10}{0.32} & \cellcolor{gray!10}{0.03} & \cellcolor{gray!10}{0.62} & \cellcolor{gray!10}{0.04} & \cellcolor{gray!10}{-6.14} & \cellcolor{gray!10}{$<$\textbf{0.01}}\\
High School Math & Llama-3.1-8b & AI letter zero shot vs. UserAI answer first & 0.32 & 0.03 & 0.64 & 0.04 & -6.89 & $<$\textbf{0.01}\\
\addlinespace
\cellcolor{gray!10}{High School Math} & \cellcolor{gray!10}{Llama-3.1-8b} & \cellcolor{gray!10}{AI letter few shot vs. UserAI direct to ai} & \cellcolor{gray!10}{0.30} & \cellcolor{gray!10}{0.02} & \cellcolor{gray!10}{0.62} & \cellcolor{gray!10}{0.04} & \cellcolor{gray!10}{-7.09} & \cellcolor{gray!10}{$<$\textbf{0.01}}\\
High School Math & Llama-3.1-8b & AI letter few shot vs. UserAI answer first & 0.30 & 0.02 & 0.64 & 0.04 & -7.98 & $<$\textbf{0.01}\\
\cellcolor{gray!10}{High School Math} & \cellcolor{gray!10}{Llama-3.1-8b} & \cellcolor{gray!10}{AI free text vs. UserAI direct to ai} & \cellcolor{gray!10}{0.64} & \cellcolor{gray!10}{0.04} & \cellcolor{gray!10}{0.62} & \cellcolor{gray!10}{0.04} & \cellcolor{gray!10}{0.24} & \cellcolor{gray!10}{0.81}\\
High School Math & Llama-3.1-8b & AI free text vs. UserAI answer first & 0.64 & 0.04 & 0.64 & 0.04 & -0.16 & 0.87\\
\cellcolor{gray!10}{High School Math} & \cellcolor{gray!10}{Llama-3.1-8b} & \cellcolor{gray!10}{User alone vs. UserAI direct to ai} & \cellcolor{gray!10}{0.45} & \cellcolor{gray!10}{0.03} & \cellcolor{gray!10}{0.62} & \cellcolor{gray!10}{0.04} & \cellcolor{gray!10}{-3.37} & \cellcolor{gray!10}{$<$\textbf{0.01}}\\
\addlinespace
High School Math & Llama-3.1-8b & User alone vs. UserAI answer first & 0.45 & 0.03 & 0.64 & 0.04 & -3.93 & $<$\textbf{0.01}\\
\cellcolor{gray!10}{College Math} & \cellcolor{gray!10}{Llama-3.1-8b} & \cellcolor{gray!10}{AI letter zero shot vs. UserAI direct to ai} & \cellcolor{gray!10}{0.35} & \cellcolor{gray!10}{0.04} & \cellcolor{gray!10}{0.46} & \cellcolor{gray!10}{0.07} & \cellcolor{gray!10}{-1.37} & \cellcolor{gray!10}{0.17}\\
College Math & Llama-3.1-8b & AI letter zero shot vs. UserAI answer first & 0.35 & 0.04 & 0.48 & 0.07 & -1.56 & 0.12\\
\cellcolor{gray!10}{College Math} & \cellcolor{gray!10}{Llama-3.1-8b} & \cellcolor{gray!10}{AI letter few shot vs. UserAI direct to ai} & \cellcolor{gray!10}{0.30} & \cellcolor{gray!10}{0.04} & \cellcolor{gray!10}{0.46} & \cellcolor{gray!10}{0.07} & \cellcolor{gray!10}{-1.97} & \cellcolor{gray!10}{\textbf{0.05}}\\
College Math & Llama-3.1-8b & AI letter few shot vs. UserAI answer first & 0.30 & 0.04 & 0.48 & 0.07 & -2.18 & \textbf{0.03}\\
\addlinespace
\cellcolor{gray!10}{College Math} & \cellcolor{gray!10}{Llama-3.1-8b} & \cellcolor{gray!10}{AI free text vs. UserAI direct to ai} & \cellcolor{gray!10}{0.41} & \cellcolor{gray!10}{0.05} & \cellcolor{gray!10}{0.46} & \cellcolor{gray!10}{0.07} & \cellcolor{gray!10}{-0.57} & \cellcolor{gray!10}{0.57}\\
College Math & Llama-3.1-8b & AI free text vs. UserAI answer first & 0.41 & 0.05 & 0.48 & 0.07 & -0.74 & 0.46\\
\cellcolor{gray!10}{College Math} & \cellcolor{gray!10}{Llama-3.1-8b} & \cellcolor{gray!10}{User alone vs. UserAI direct to ai} & \cellcolor{gray!10}{0.40} & \cellcolor{gray!10}{0.04} & \cellcolor{gray!10}{0.46} & \cellcolor{gray!10}{0.07} & \cellcolor{gray!10}{-0.75} & \cellcolor{gray!10}{0.46}\\
College Math & Llama-3.1-8b & User alone vs. UserAI answer first & 0.40 & 0.04 & 0.48 & 0.07 & -0.93 & 0.35\\
\cellcolor{gray!10}{Conceptual Physics} & \cellcolor{gray!10}{Llama-3.1-8b} & \cellcolor{gray!10}{AI letter zero shot vs. UserAI direct to ai} & \cellcolor{gray!10}{0.53} & \cellcolor{gray!10}{0.05} & \cellcolor{gray!10}{0.67} & \cellcolor{gray!10}{0.04} & \cellcolor{gray!10}{-2.25} & \cellcolor{gray!10}{\textbf{0.02}}\\
\addlinespace
Conceptual Physics & Llama-3.1-8b & AI letter zero shot vs. UserAI answer first & 0.53 & 0.05 & 0.73 & 0.04 & -3.22 & $<$\textbf{0.01}\\
\cellcolor{gray!10}{Conceptual Physics} & \cellcolor{gray!10}{Llama-3.1-8b} & \cellcolor{gray!10}{AI letter few shot vs. UserAI direct to ai} & \cellcolor{gray!10}{0.57} & \cellcolor{gray!10}{0.04} & \cellcolor{gray!10}{0.67} & \cellcolor{gray!10}{0.04} & \cellcolor{gray!10}{-1.64} & \cellcolor{gray!10}{0.10}\\
Conceptual Physics & Llama-3.1-8b & AI letter few shot vs. UserAI answer first & 0.57 & 0.04 & 0.73 & 0.04 & -2.70 & $<$\textbf{0.01}\\
\cellcolor{gray!10}{Conceptual Physics} & \cellcolor{gray!10}{Llama-3.1-8b} & \cellcolor{gray!10}{AI free text vs. UserAI direct to ai} & \cellcolor{gray!10}{0.62} & \cellcolor{gray!10}{0.04} & \cellcolor{gray!10}{0.67} & \cellcolor{gray!10}{0.04} & \cellcolor{gray!10}{-0.77} & \cellcolor{gray!10}{0.44}\\
Conceptual Physics & Llama-3.1-8b & AI free text vs. UserAI answer first & 0.62 & 0.04 & 0.73 & 0.04 & -1.80 & 0.07\\
\addlinespace
\cellcolor{gray!10}{Conceptual Physics} & \cellcolor{gray!10}{Llama-3.1-8b} & \cellcolor{gray!10}{User alone vs. UserAI direct to ai} & \cellcolor{gray!10}{0.46} & \cellcolor{gray!10}{0.03} & \cellcolor{gray!10}{0.67} & \cellcolor{gray!10}{0.04} & \cellcolor{gray!10}{-3.91} & \cellcolor{gray!10}{$<$\textbf{0.01}}\\
Conceptual Physics & Llama-3.1-8b & User alone vs. UserAI answer first & 0.46 & 0.03 & 0.73 & 0.04 & -4.97 & $<$\textbf{0.01}\\
\cellcolor{gray!10}{Moral Scenarios} & \cellcolor{gray!10}{Llama-3.1-8b} & \cellcolor{gray!10}{AI letter zero shot vs. UserAI direct to ai} & \cellcolor{gray!10}{0.40} & \cellcolor{gray!10}{0.03} & \cellcolor{gray!10}{0.72} & \cellcolor{gray!10}{0.04} & \cellcolor{gray!10}{-6.01} & \cellcolor{gray!10}{$<$\textbf{0.01}}\\
Moral Scenarios & Llama-3.1-8b & AI letter zero shot vs. UserAI answer first & 0.40 & 0.03 & 0.74 & 0.04 & -7.42 & $<$\textbf{0.01}\\
\cellcolor{gray!10}{Moral Scenarios} & \cellcolor{gray!10}{Llama-3.1-8b} & \cellcolor{gray!10}{AI letter few shot vs. UserAI direct to ai} & \cellcolor{gray!10}{0.31} & \cellcolor{gray!10}{0.03} & \cellcolor{gray!10}{0.72} & \cellcolor{gray!10}{0.04} & \cellcolor{gray!10}{-7.35} & \cellcolor{gray!10}{$<$\textbf{0.01}}\\
\addlinespace
Moral Scenarios & Llama-3.1-8b & AI letter few shot vs. UserAI answer first & 0.31 & 0.03 & 0.74 & 0.04 & -8.86 & $<$\textbf{0.01}\\
\cellcolor{gray!10}{Moral Scenarios} & \cellcolor{gray!10}{Llama-3.1-8b} & \cellcolor{gray!10}{AI free text vs. UserAI direct to ai} & \cellcolor{gray!10}{0.49} & \cellcolor{gray!10}{0.03} & \cellcolor{gray!10}{0.72} & \cellcolor{gray!10}{0.04} & \cellcolor{gray!10}{-4.07} & \cellcolor{gray!10}{$<$\textbf{0.01}}\\
Moral Scenarios & Llama-3.1-8b & AI free text vs. UserAI answer first & 0.49 & 0.03 & 0.74 & 0.04 & -5.15 & $<$\textbf{0.01}\\
\cellcolor{gray!10}{Moral Scenarios} & \cellcolor{gray!10}{Llama-3.1-8b} & \cellcolor{gray!10}{User alone vs. UserAI direct to ai} & \cellcolor{gray!10}{0.79} & \cellcolor{gray!10}{0.03} & \cellcolor{gray!10}{0.72} & \cellcolor{gray!10}{0.04} & \cellcolor{gray!10}{1.34} & \cellcolor{gray!10}{0.18}\\
Moral Scenarios & Llama-3.1-8b & User alone vs. UserAI answer first & 0.79 & 0.03 & 0.74 & 0.04 & 1.00 & 0.32\\
\bottomrule
\end{tabular}
\caption{Results per dataset for Llama-3.1-8b, including AI-alone vs. user-AI comparisons and user-alone vs. user-AI comparisons.}
\label{tab:llama-results}
\end{adjustwidth}
\end{table*}

%% file: tables/chatbench_stats.tex
\begin{table*}
\centering
\small
\begin{tabular}{llllr}
\toprule
 &  &  &  & Count \\
Model & Dataset & Condition & Answer Type &  \\
\midrule
\multirow[t]{20}{*}{GPT-4o} & \multirow[t]{4}{*}{College Math} & \multirow[t]{2}{*}{answer-first} & userAIAnswer & 134 \\
 &  &  & userAnswer & 283 \\
\cline{3-5}
 &  & \multirow[t]{2}{*}{direct-to-AI} & userAIAnswer & 116 \\
 &  &  & userAnswer & 121 \\
\cline{2-5} \cline{3-5}
 & \multirow[t]{4}{*}{Conceptual Physics} & \multirow[t]{2}{*}{answer-first} & userAIAnswer & 317 \\
 &  &  & userAnswer & 425 \\
\cline{3-5}
 &  & \multirow[t]{2}{*}{direct-to-AI} & userAIAnswer & 351 \\
 &  &  & userAnswer & 117 \\
\cline{2-5} \cline{3-5}
 & \multirow[t]{4}{*}{Elementary Math} & \multirow[t]{2}{*}{answer-first} & userAIAnswer & 542 \\
 &  &  & userAnswer & 697 \\
\cline{3-5}
 &  & \multirow[t]{2}{*}{direct-to-AI} & userAIAnswer & 462 \\
 &  &  & userAnswer & 122 \\
\cline{2-5} \cline{3-5}
 & \multirow[t]{4}{*}{High School Math} & \multirow[t]{2}{*}{answer-first} & userAIAnswer & 539 \\
 &  &  & userAnswer & 689 \\
\cline{3-5}
 &  & \multirow[t]{2}{*}{direct-to-AI} & userAIAnswer & 463 \\
 &  &  & userAnswer & 122 \\
\cline{2-5} \cline{3-5}
 & \multirow[t]{4}{*}{Moral Scenarios} & \multirow[t]{2}{*}{answer-first} & userAIAnswer & 242 \\
 &  &  & userAnswer & 331 \\
\cline{3-5}
 &  & \multirow[t]{2}{*}{direct-to-AI} & userAIAnswer & 398 \\
 &  &  & userAnswer & 135 \\
\cline{1-5} \cline{2-5} \cline{3-5}
\multirow[t]{20}{*}{Llama-3.1-8b} & \multirow[t]{4}{*}{College Math} & \multirow[t]{2}{*}{answer-first} & userAIAnswer & 119 \\
 &  &  & userAnswer & 251 \\
\cline{3-5}
 &  & \multirow[t]{2}{*}{direct-to-AI} & userAIAnswer & 115 \\
 &  &  & userAnswer & 123 \\
\cline{2-5} \cline{3-5}
 & \multirow[t]{4}{*}{Conceptual Physics} & \multirow[t]{2}{*}{answer-first} & userAIAnswer & 315 \\
 &  &  & userAnswer & 428 \\
\cline{3-5}
 &  & \multirow[t]{2}{*}{direct-to-AI} & userAIAnswer & 333 \\
 &  &  & userAnswer & 112 \\
\cline{2-5} \cline{3-5}
 & \multirow[t]{4}{*}{Elementary Math} & \multirow[t]{2}{*}{answer-first} & userAIAnswer & 485 \\
 &  &  & userAnswer & 620 \\
\cline{3-5}
 &  & \multirow[t]{2}{*}{direct-to-AI} & userAIAnswer & 462 \\
 &  &  & userAnswer & 123 \\
\cline{2-5} \cline{3-5}
 & \multirow[t]{4}{*}{High School Math} & \multirow[t]{2}{*}{answer-first} & userAIAnswer & 477 \\
 &  &  & userAnswer & 610 \\
\cline{3-5}
 &  & \multirow[t]{2}{*}{direct-to-AI} & userAIAnswer & 464 \\
 &  &  & userAnswer & 125 \\
\cline{2-5} \cline{3-5}
 & \multirow[t]{4}{*}{Moral Scenarios} & \multirow[t]{2}{*}{answer-first} & userAIAnswer & 349 \\
 &  &  & userAnswer & 471 \\
\cline{3-5}
 &  & \multirow[t]{2}{*}{direct-to-AI} & userAIAnswer & 229 \\
 &  &  & userAnswer & 81 \\
\cline{1-5} \cline{2-5} \cline{3-5}
\bottomrule
\end{tabular}
\caption{Dataset statistics for ChatBench.}
\label{tab:data_stats}
\end{table*}

%% file: app_analysis.tex
\section{Details on Analyses and Experiments}
\label{sec:app_analyses}

We download the MMLU datasets\footnote{\url{https://huggingface.co/datasets/cais/mmlu}.} \cite{hendrycks2021mmlu} and MMLU-Redux  datasets\footnote{\url{https://huggingface.co/datasets/edinburgh-dawg/mmlu-redux-2.0}.} \cite{gema2024mmluredux} from Hugging Face.
The datasets are protected by the MIT and CC-by-4.0 licenses, respectively, allowing our use of this data in our research.
Our code is available at \gitrepo{}.

\subsection{AI-Alone experiments}
\label{sec:ai_alone_experiments}
Here we provide the exact prompts used for each of the AI-alone methods: few-shot letter-only (Listing \ref{lst:few_shot_prompt}), zero-shot letter-only (Listing \ref{lst:zero_shot_prompt}), and the two prompts for free-text (Listing \ref{lst:free_text_prompt_1} and Listing \ref{lst:free_text_prompt_2}).
For all the methods, the system prompt was ``You are a helpful AI assistant.''

\lstset{
  breaklines=true,     
  frame=single,        
  basicstyle=\tiny\ttfamily, 
  showstringspaces=false  
}
\begin{lstlisting}[caption={Prompt for few-shot letter-only, taken from HELM. In-context examples are the five examples in MMLU's ``dev'' set for this dataset.},label={lst:few_shot_prompt}]
Answer with only a single letter.

The following are multiple choice questions (with answers) about {dataset}.

{example_1}
A. {example_1_option_A}
B. {example_1_option_B}
C. {example_1_option_C}
D. {example_1_option_D}
Answer: {example_1_answer}

...

{example_5}
A. {example_5_option_A}
B. {example_5_option_B}
C. {example_5_option_C}
D. {example_5_option_D}
Answer: {example_5_answer}

{question}
A. {option_A}
B. {option_B}
C. {option_C}
D. {option_D}
Answer:
\end{lstlisting}

\begin{lstlisting}[caption={Prompt for zero-shot letter-only, using the same language as few-shot but dropping the in-context examples.},label={lst:zero_shot_prompt}]
Answer with only a single letter.

{question}
A. {option_A}
B. {option_B}
C. {option_C}
D. {option_D}
Answer:
\end{lstlisting}

\begin{lstlisting}[caption={First prompt for AI-alone free-text. This prompt to generate the model's free-text response is simply the question and answer options concatenated.},label={lst:free_text_prompt_1}]
{question}
A. {option_A}
B. {option_B}
C. {option_C}
D. {option_D}
\end{lstlisting}

\begin{lstlisting}[caption={Second prompt for AI-alone free-text. This second prompt instructs GPT-4o to extract an answer (if any) from the model's free-text response. In order to not bias the answer extraction, we do not include the correct answer in this prompt.},label={lst:free_text_prompt_2}]
Here is a question that someone was asked:

================================================
{question}
A. {option_A}
B. {option_B}
C. {option_C}
D. {option_D}
================================================

Here is a response:

================================================
{response}
================================================

Did the response provide a final answer to the question?Respond with a JSON object that contains one key "attempted_answer" with a value that is true or false. If "attempted_answer" is true, then include a second key "answer_val" with the final answer's value in quotations. If the final answer value matches one of the answer options, include a third key "answer_letter" with a value that is one of the letters "A", "B", "C", or "D".
\end{lstlisting}

In \dataname{}, we include the results of our AI-alone experiments, where we tested each of the two models (GPT-4o and Llama-3.1-8b) 50 times per question and AI-alone method.
Testing 50 times was necessary since we used a temperature of 0.7, as discussed in the main text.
We were able to get 50 answers for almost every model, question, and method, barring a few exceptions.
For the letter-only methods, the model would occasionally not return a valid answer, since its response would begin with a character besides ``A'', ``B'', ``C'', or ``D''. 
Thus, we computed two accuracies: one where the invalid answers were treated as incorrect (since the model failed to follow instructions) and one where we computed accuracy over only the valid answers.
We report the former accuracy in the paper, but report both types of accuracies in \dataname{}.
As expected, invalid answers were more common with zero-shot than few-shot, but they were a minor occurrence overall: below 5\% of answers were invalid for 90\% of questions with zero-shot and 99.6\% of questions with few-shot. 
Invalid answers were not an issue for free-text, but we very occasionally had issues with the answer extraction step (e.g., errors in JSON parsing), resulting in losing one or two out of 50 answers for a few questions.
For one of the Moral Scenarios questions, we had issues generating free-text responses, since it violated our model deployment's filtering policy.

\subsection{Statistical details}
\label{sec:statistical_tests}

\paragraph{Mean accuracies.}
When measuring accuracies for all methods (user-alone, AI-alone, and user-AI), we first compute per-question accuracies as the fraction of correct answers over total answers $n_q$ for each question, denoted $\hat{p}_q$. We also compute the standard error for each question-level accuracy estimate $SE_q = \sqrt{\hat{p}_q(1-\hat{p}_q) / n_q}$. We then compute dataset-level accuracies with an (unweighted) average across all $Q$ question-level accuracies, and dataset-level standard errors using decomposition of total variance to account for both variability in sampling questions from the larger population of MMLU questions and variability in correctness of responses \cite{miller2024adding}: 
\begin{align}
    SE_{\mathrm{tot}} = \sqrt{(\mathrm{E}[SE_q] + \mathrm{Var}(\hat{p}_q)) / Q}.
\end{align}

In Tables \ref{tab:gpt4o-results} and \ref{tab:llama-results}, we report mean accuracies for all datasets, models, AI-alone methods, and user-AI conditions. 
We also compare accuracies between two methods, for AI-alone vs. user-AI and for user-alone vs. user-AI.
We conduct z-tests for all statistical tests comparing accuracies between two methods where 
\begin{align}
    z = (\hat{p}_1 - \hat{p}_2) / \sqrt{SE_1^2 + SE_2^2}.
\end{align}

\paragraph{Upper-bound on correlation.}
Since there is noise in our estimate of user-AI accuracy per question, we want to check if the low correlations between user-AI and AI-alone accuracies can be explained by that noise.
To test this, we simulate an upper bound on what the correlation would be if the user-AI accuracies were drawn from the same distribution as the AI-alone accuracies, which we assume are perfectly estimated because we test each LLM 50 times on each question.
We construct hypothetical user-AI data, where for each question $q$, we draw $x$ from Binomial($n_{qu}$, $p_{qf}$), where $n_{qu}$ is the number of user-AI answers we had in our study, $p_{qf}$ is the free-text accuracy on this question, and $x/n_{qu}$ is now the hypothetical user-AI accuracy on this question.
Over 100 iterations, the hypothetical correlation with free-text ranges from $r=0.88$ to $0.92$ for \textit{direct-to-AI} and $r=0.90$ to $0.94$ for \textit{answer-first}, with the real correlations of $0.45$ and $0.46$ (respectively) well below this range.

\subsection{Additional analyses from user study}
\paragraph{User-AI conversations.}
To characterize the conversations between users and LLMs, we set a system prompt of ``You are an expert at extracting information from conversations and MUST return a JSON object.'' and used the prompt in Listing \ref{lst:conversation_prompt}. 
\begin{lstlisting}[caption={Prompt to GPT-4o for automatically characterizing user-AI conversations.},label={lst:conversation_prompt}]
The following conversation occurred between a user called "You" and an AI called "Bot", in which the user "You" was given a question and tried to use "Bot" to answer it correctly. Answer the following queries about this conversation.

    1. True or false: the first user question is an exact or near-exact rephrasing of the question that the user was given, to which an intelligent person or AI would respond with the correct answer among the answer choices (without seeing anything else, including the answer choices if the user didn't supply them). Answer with "true" or "false".
    2. True or false: the first time the AI answers the question, it is with a correct answer to the question that the user was given. Answer with "true" or "false".
    3. True or false: the AI answers the question that the user was given more than once during the conversation. Answer with "true" or "false".
    4: The first time the AI answers the question, which answer choice does its answer correspond to? Answer with "A", "B", "C", "D", or "none".

    Here is the question the user was given:
    {question}

    {choices}

    Here is the correct answer to the question:
    {correctAnswer}

    Here is the conversation between the user ("You") and the AI ("Bot"):
    {conversation}

    Format your answer as a JSON object with the following keys: "1:first_user_question","2:first_ai_answer_correct","3:more_than_one_ai_answer","4:first_ai_answer_option".
\end{lstlisting}

\begin{figure}[t]
    \centering
    \includegraphics[width=\linewidth]{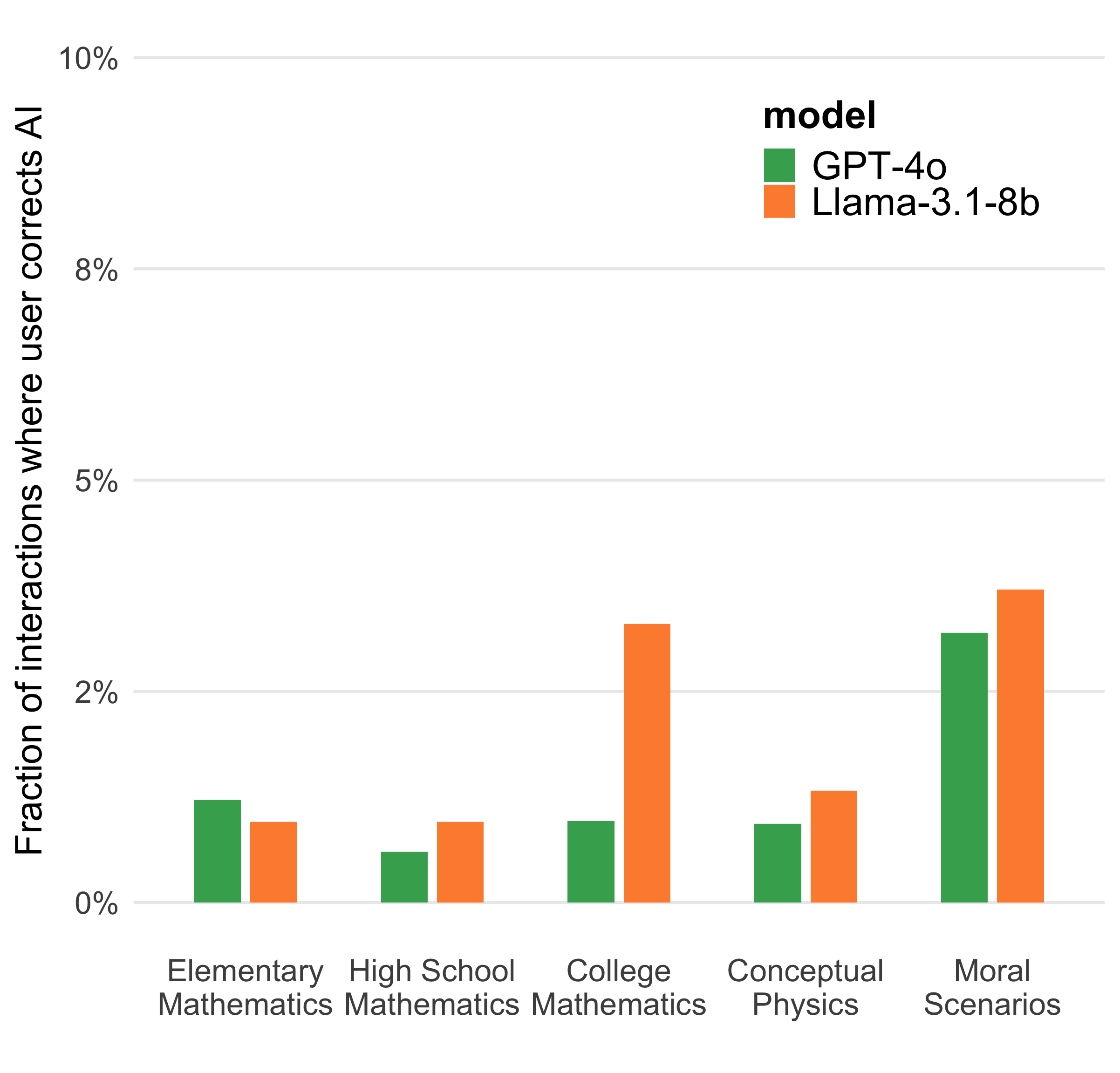}
    \caption{Fraction of user-AI interactions where the last AI answer in the conversation is wrong but the user still answered correctly, by subject and model.}
    \label{fig:user_corrects_ai}
\end{figure}

We used the structured data from this analysis for a number of results, described in Section~\ref{sec:conversations}, such as how often user-AI interactions ``mirror'' AI benchmarks and how often AI provided the correct answer in the user-AI conversations.
We also used this structured data to measure how often the user corrects the AI model's mistake, by computing the fraction of user-AI interactions where the first AI answer in the conversation is wrong but the user still answered correctly (Figure \ref{fig:user_corrects_ai}).
We find that this occurs in a small fraction of cases, and users are generally more likely to correct Llama-3.1-8b than GPT-4o. 

In Table \ref{tab:conversation_stats}, we provide additional statistics of the user-AI conversations, some computed from the structured data described above.
Overall, we find that the conversations tend to be short in length, but each message is long and there is genuine information being exchanged. This is natural in a QA setting where we are only asking the user to answer one single question (no follow-ups) with the help of AI. Even without those constraints, user-AI conversations in the real world are also often short: for example, in WildChat \cite{zhao2024wildchat}, which includes 1 million user-AI conversations “in the wild”, around 60\% of their conversations consist of only one user-AI turn. 

In \dataname{}, we also find that the vast majority of users are putting in effort to try to use AI to answer the question, as opposed to not putting in effort (e.g., “hello”, without a second message) or not using the AI (e.g., “i solved this already”). 
Oftentimes, this effort takes the form of the user providing a near-exact rephrasing of the question or a component of the question. The user’s effort is substantiated by what follows, since in the majority of conversations, we see AI providing an explicit answer to the question. We also find that the user is clearly adapting their answers based on the AI: the rate that user-AI is right is much higher when AI is right, than when AI is wrong or when the user is answering alone. 
Finally, we also see interesting differences across subjects: Moral Scenarios is distinguished by greater user independence and less reliance on AI, resulting in fewer near-exact rephrasings and explicit AI answers, and greater divergence between the AI answer being right and the user-AI answer being right.

\begin{figure*}[t]
    \centering
    \includegraphics[width=0.8\linewidth]{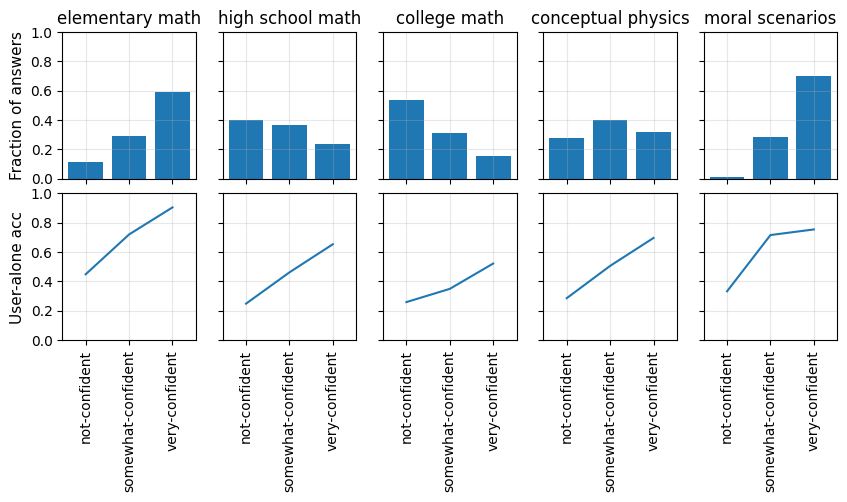}
    \caption{Distribution of confidence answers from users and mean user-alone accuracies per confidence answer.}
    \label{fig:confidence_plot}
\end{figure*}
\paragraph{User confidence.} 
In Figure \ref{fig:confidence_plot}, we visualize the relationship between user-reported confidence per question and their user-alone accuracy. 
First, over our five datasets, we find that users are most confident about Moral Scenarios, followed by Elementary Math, Conceptual Physics, High School Math, and College Math.
The user selects their confidence from three options (as shown Figure \ref{fig:confidence}), ``not confident'', ``somewhat confident'', and ``very confident''. 
We find that users are well-calibrated within dataset: as their confidence increases, so does the mean accuracy.
Users are less well-calibrated across datasets: for example, users who are very confident on a Conceptual Physics question slightly underperform those who are only somewhat confident on an Elementary Mathematics question. 

\include{tables/conversation_stats}

\subsection{Simulator details}
Below we provide the exact prompts for the two-step simulator (Listings \ref{lst:two_step_sys_prompt}-\ref{lst:two_step_user_prompt_2}) and the IQA-EVAL simulator from \citet{li2024iqaeval} (Listing \ref{lst:iqa_eval_prompt}).
\begin{lstlisting}[caption={Two-step user simulator, system prompt for both tasks.},label={lst:two_step_sys_prompt}]
You are a human user interacting with an AI system, and you are trying to answer the following question:

{question}
A. {option_A}
B. {option_B}
C. {option_C}
D. {option_D}
\end{lstlisting}

\begin{lstlisting}[caption={Two-step user simulator, user prompt for Task 1 (user refers to the role in the OpenAI API, not a real user).},label={lst:two_step_user_prompt_1}]
Generate the first prompt you would say to the system to get started with answering your question. Remember to write exactly as a real user would.
\end{lstlisting}

\begin{lstlisting}[caption={Two-step user simulator, user prompt for Task 2 (user refers to the role in the OpenAI API, not a real user).},label={lst:two_step_user_prompt_2}]
Here is your conversation so far with the AI system:
========================
YOU: {simulator prompt 1}

SYSTEM: {AI system response 1}

...

YOU: {simluator prompt k}

SYSTEM: {AI system response k}
========================
If your question is answered by this conversation, return ONLY the answer in the format "Answer: A, B, C, or D". If not, generate the next prompt you would say to the system to answer your question. Remember to keep your writing style consistent.
\end{lstlisting}

\begin{lstlisting}[caption={IQA-EVAL simulator, only has system prompt, following the original implementation.},label={lst:iqa_eval_prompt}]
You are mimicking a human.
You are trying to choose the correct answer to the given question.
Please ask an assistant sub-questions for help approaching answers.
In each turn, please only ask one sub-question to interact with an assistant. In the sub-questions, please include all necessary information, such as the question and options, in the original question. If you know the answer, please output "So, the answer is: A, B, C, or D."
{question}
A. {option_A}
B. {option_B}
C. {option_C}
D. {option_D}

YOU: {simulator prompt 1}

SYSTEM: {AI system response 1}

...

YOU: {simluator prompt k}

SYSTEM: {AI system response k}
\end{lstlisting}

\begin{figure*}
    \centering
    \includegraphics[width=\linewidth]{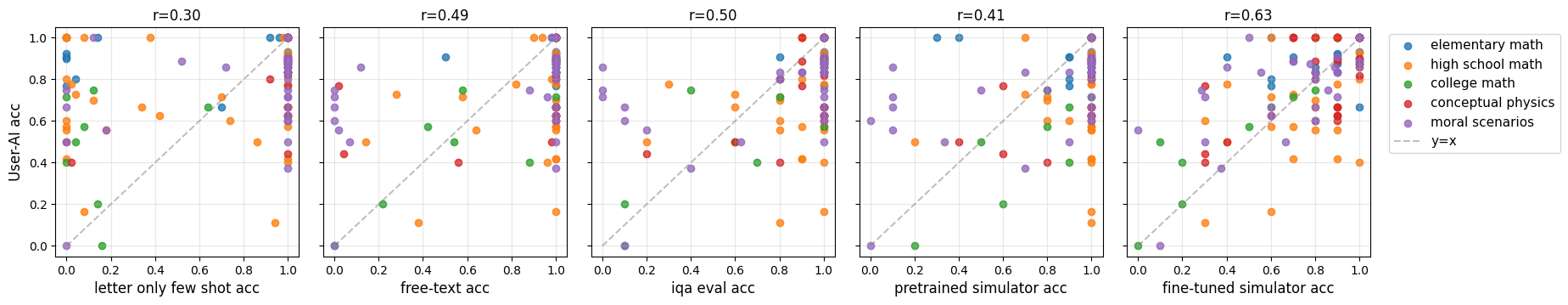}
    \caption{Scatter plot comparing different AI-alone and user simulator methods' abilities to predict user-AI accuracy, where the AI system is GPT-4o. Pearson correlations are included in the plot titles.}
    \label{fig:gpt4o-sim}
\end{figure*}

\begin{figure*}
    \centering
    \includegraphics[width=\linewidth]{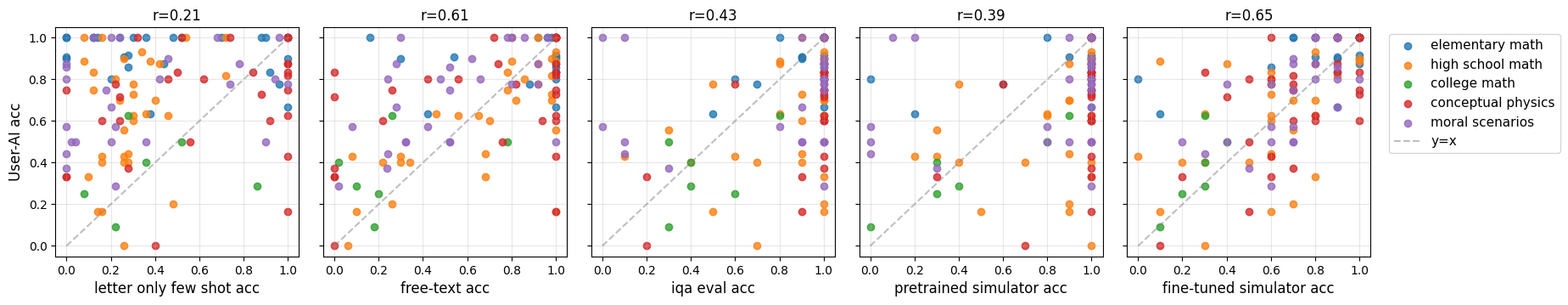}
    \caption{Scatter plot comparing different AI-alone and user simulator methods' abilities to predict user-AI accuracy, where the AI system is Llama-3.1-8b. Pearson correlations are included in the plot titles.}
    \label{fig:llama-sim}
\end{figure*}

In our simulator experiments, we fine-tune GPT-4o using Azure OpenAI Service.
We use the default hyperparameters, with a batch size of 11 and 2 epochs.
The training data contains 8,538 training examples (we describe in Section \ref{sec:simulator} how each user-AI conversation with $k$ user utterances becomes $k+1$ training examples for fine-tuning).

%% file: tables/conversation_stats.tex
\begin{table*}[t]
\centering
\begin{tabular}{lccc}
\toprule
\textbf{Statistic} & \textbf{Math} & \textbf{Conceptual Physics} & \textbf{Moral Scenarios} \\
\midrule
Avg \# user messages & 1.57 & 1.37 & 1.69 \\
Avg character length of user message & 88.21 & 72.93 & 105.24 \\
Avg \# AI messages & 1.48 & 1.32 & 1.51 \\
Avg character length of AI message & 793.92 & 918.40 & 1073.03 \\
\midrule
Conversation length: 2+ messages (\%) & 98\% & 98\% & 97\% \\
Conversation length: 4+ messages (\%) & 27\% & 21\% & 37\% \\
Conversation length: 6+ messages (\%) & 11\% & 8\% & 11\% \\
\midrule
First prompt: near-exact rephrase (\%) & 58\% & 62\% & 23\% \\
First prompt: component (\%) & 25\% & 29\% & 63\% \\
Explicit AI answer in conversation (\%) & 86\% & 82\% & 50\% \\
\midrule
AI answer right (\%) & 76\% & 84\% & 78\% \\
AI answer right, user-AI right (\%) & 91\% & 91\% & 82\% \\
AI answer wrong, user-AI right (\%) & 42\% & 32\% & 56\% \\
\bottomrule
\end{tabular}
\caption{Statistics of user-AI conversations in \dataname{}, computed over the full study.}
\label{tab:conversation_stats}
\end{table*}

%% file: main.bbl
\begin{thebibliography}{44}
\providecommand{\natexlab}[1]{#1}

\bibitem[{Argyle et~al.(2023)Argyle, Busby, Fulda, Gubler, Rytting, and Wingate}]{argyle2023samples}
Lisa~P. Argyle, Ethan~C. Busby, Nancy Fulda, Joshua~R. Gubler, Christopher Rytting, and David Wingate. 2023.
\newblock Out of one, many: Using language models to simulate human samples.
\newblock \emph{Political Analysis}, 31:337--351.

\bibitem[{Bansal et~al.(2021)Bansal, Wu, Zhou, Fok, Nushi, Kamar, Ribeiro, and Weld}]{bansal2021whole}
Gagan Bansal, Tongshuang Wu, Joyce Zhou, Raymond Fok, Besmira Nushi, Ece Kamar, Marco~Tulio Ribeiro, and Daniel Weld. 2021.
\newblock Does the whole exceed its parts? the effect of ai explanations on complementary team performance.
\newblock In \emph{CHI '21: Proceedings of the 2021 CHI Conference on Human Factors in Computing Systems}, 81, pages 1--16.

\bibitem[{Bick et~al.(2024)Bick, Blandin, and Deming}]{bick2024adoption}
Alexander Bick, Adam Blandin, and David Deming. 2024.
\newblock The rapid adoption of generative {AI}.
\newblock \emph{Federal Reserve Bank of St. Louis}.
\newblock \url{https://www.stlouisfed.org/on-the-economy/2024/sep/rapid-adoption-generative-ai}.

\bibitem[{Bisbee et~al.(2024)Bisbee, Clinton, Dorff, Kenkel, and Larson}]{bisbee2024perils}
James Bisbee, Joshua~D. Clinton, Cassy Dorff, Brenton Kenkel, and Jennifer~M. Larson. 2024.
\newblock Synthetic replacements for human survey data? the perils of large language models.
\newblock \emph{Political Analysis}, 32:401--416.

\bibitem[{Chang et~al.(2024)Chang, Chaszczewicz, Wang, Josifovska, Pierson, and Leskovec}]{chang2024network}
Serina Chang, Alicja Chaszczewicz, Emma Wang, Maya Josifovska, Emma Pierson, and Jure Leskovec. 2024.
\newblock {LLMs} generate structurally realistic social networks but overestimate political homophily.
\newblock \emph{arXiv preprint arXiv:2408.16629}.

\bibitem[{Cheng et~al.(2023{\natexlab{a}})Cheng, Durmus, and Jurafsky}]{cheng2023marked}
Myra Cheng, Esin Durmus, and Dan Jurafsky. 2023{\natexlab{a}}.
\newblock Marked personas: Using natural language prompts to measure stereotypes in language models.
\newblock In \emph{Proceedings of the 61st Annual Meeting of the Association for Computational Linguistics (ACL'23)}.

\bibitem[{Cheng et~al.(2023{\natexlab{b}})Cheng, Piccardi, and Yang}]{cheng2023compost}
Myra Cheng, Tiziano Piccardi, and Diyi Yang. 2023{\natexlab{b}}.
\newblock Compost: Characterizing and evaluating caricature in {LLM} simulations.
\newblock In \emph{Proceedings of the 2023 Conference on Empirical Methods in Natural Language Processing (EMNLP'23)}.

\bibitem[{Chiang et~al.(2024)Chiang, Zheng, Sheng, Angelopoulos, Li, Li, Zhang, Zhu, Jordan, Gonzalez, and Stoica}]{chiang2024chatbotarena}
Wei-Lin Chiang, Lianmin Zheng, Ying Sheng, Anastasios~Nikolas Angelopoulos, Tianle Li, Dacheng Li, Hao Zhang, Banghua Zhu, Michael Jordan, Joseph~E. Gonzalez, and Ion Stoica. 2024.
\newblock {Chatbot Arena}: An open platform for evaluating {LLMs} by human preference.
\newblock \emph{arXiv preprint arXiv:2403.04132}.

\bibitem[{Cobbe et~al.(2021)Cobbe, Kosaraju, Bavarian, Chen, Jun, Kaiser, Plappert, Tworek, Hilton, Nakano, Hesse, and Schulman}]{cobbe2021gsm8k}
Karl Cobbe, Vineet Kosaraju, Mohammad Bavarian, Mark Chen, Heewoo Jun, Lukasz Kaiser, Matthias Plappert, Jerry Tworek, Jacob Hilton, Reiichiro Nakano, Christopher Hesse, and John Schulman. 2021.
\newblock Training verifiers to solve math word problems.
\newblock \emph{arXiv preprint arXiv:2110.14168}.

\bibitem[{Collins et~al.(2024)Collins, Jiang, Frieder, Wong, Zilka, Bhatt, Lukasiewicz, Wu, Tenenbaum, Hart, Gowers, Li, Weller, and Jamnik}]{collins2024theorem}
Katherine~M. Collins, Albert~Q. Jiang, Simon Frieder, Lionel Wong, Miri Zilka, Umang Bhatt, Thomas Lukasiewicz, Yuhuai Wu, Joshua~B. Tenenbaum, William Hart, Timothy Gowers, Wenda Li, Adrian Weller, and Mateja Jamnik. 2024.
\newblock Evaluating language models for mathematics through interactions.
\newblock \emph{Proceedings of the National Academy of Sciences (PNAS)}.

\bibitem[{Dhillon et~al.(2024)Dhillon, Molaei, Li, Golub, Zheng, and Robert}]{dhillon2024cowriting}
Paramveer~S. Dhillon, Somayeh Molaei, Jiaqi Li, Maximilian Golub, Shaochun Zheng, and Lionel~Peter Robert. 2024.
\newblock Shaping human-ai collaboration: Varied scaffolding levels in co-writing with language models.
\newblock In \emph{Proceedings of the 2024 CHI Conference on Human Factors in Computing Systems (CHI'24)}.

\bibitem[{Dubois et~al.(2023)Dubois, Li, Taori, Zhang, Gulrajani, Ba, Guestrin, Liang, and Hashimoto}]{dubois2023alpacafarm}
Yann Dubois, Xuechen Li, Rohan Taori, Tianyi Zhang, Ishaan Gulrajani, Jimmy Ba, Carlos Guestrin, Percy Liang, and Tatsunori~B. Hashimoto. 2023.
\newblock Alpacafarm: A simulation framework for methods that learn from human feedback.
\newblock In \emph{Proceedings of the 37th International Conference on Neural Information Processing Systems (NeurIPS'23)}.

\bibitem[{Gema et~al.(2024)Gema, Leang, Hong, Devoto, Mancino, Saxena, He, Zhao, Du, Madani, Barale, McHardy, Harris, Kaddour, van Krieken, and Minervini}]{gema2024mmluredux}
Aryo~Pradipta Gema, Joshua Ong~Jun Leang, Giwon Hong, Alessio Devoto, Alberto Carlo~Maria Mancino, Rohit Saxena, Xuanli He, Yu~Zhao, Xiaotang Du, Mohammad Reza~Ghasemi Madani, Claire Barale, Robert McHardy, Joshua Harris, Jean Kaddour, Emile van Krieken, and Pasquale Minervini. 2024.
\newblock Are we done with {MMLU}?
\newblock \emph{arXiv preprint arXiv:2406.04127}.

\bibitem[{{Gemini Team Google}(2023)}]{google2023gemini}
{Gemini Team Google}. 2023.
\newblock Gemini: A family of highly capable multimodal models.
\newblock \emph{arXiv preprint arXiv:2312.11805}.

\bibitem[{Hardy et~al.(2024)Hardy, Reuel, Meimandi, Soder, Griffith, Asmar, Koyejo, Bernstein, and Kochenderfer}]{hardy2024marketing}
Amelia Hardy, Anka Reuel, Kiana~Jafari Meimandi, Lisa Soder, Allie Griffith, Dylan~M. Asmar, Sanmi Koyejo, Michael~S. Bernstein, and Mykel~J. Kochenderfer. 2024.
\newblock More than marketing? on the information value of ai benchmarks for practitioners.
\newblock \emph{arXiv preprint arXiv:2412.05520}.

\bibitem[{Hendrycks et~al.(2021)Hendrycks, Burns, Basart, Zou, Mazeika, Song, and Steinhardt}]{hendrycks2021mmlu}
Dan Hendrycks, Collin Burns, Steven Basart, Andy Zou, Mantas Mazeika, Dawn Song, and Jacob Steinhardt. 2021.
\newblock Measuring massive multitask language understanding.
\newblock In \emph{The Ninth International Conference on Learning Representations (ICLR'21)}.

\bibitem[{Hewitt et~al.(2024)Hewitt, Ashokkumar, Ghezae1, and Willer}]{hewitt2024social}
Luke Hewitt, Ashwini Ashokkumar, Isaias Ghezae1, and Robb Willer. 2024.
\newblock Predicting results of social science experiments using large language models.
\newblock \emph{Working Paper}.

\bibitem[{Horton(2023)}]{horton2023silicus}
John~J. Horton. 2023.
\newblock Large language models as simulated economic agents: What can we learn from homo silicus?
\newblock \emph{arXiv preprint arXiv:2301.07543}.

\bibitem[{Hwang et~al.(2023)Hwang, Majumder, and Tandon}]{hwang2023aligning}
EunJeong Hwang, Bodhisattwa Majumder, and Niket Tandon. 2023.
\newblock Aligning language models to user opinions.
\newblock In \emph{Findings of the Association for Computational Linguistics: EMNLP 2023}.

\bibitem[{Ipeirotis(2010)}]{ipeirotis2010turk}
Panos Ipeirotis. 2010.
\newblock Demographics of mechanical turk.
\newblock \emph{CeDER Working Papers}.

\bibitem[{Jurenka et~al.(2024)Jurenka, amd Kevin R.~McKee, Gillick, Zhu, Wiltberger, Phal, Hermann, Kasenberg, Bhoopchand et~al.}]{google2024learnlm}
Irina Jurenka, Markus~Kunesch amd Kevin R.~McKee, Daniel Gillick, Shaojian Zhu, Sara Wiltberger, Shubham~Milind Phal, Katherine Hermann, Daniel Kasenberg, Avishkar Bhoopchand, et~al. 2024.
\newblock Towards responsible development of generative ai for education: An evaluation-driven approach.
\newblock \emph{arXiv preprint arXiv:2407.12687}.

\bibitem[{Kong et~al.(2024)Kong, Fan, Wan, Jiang, and Wang}]{kong2024platolm}
Chuyi Kong, Yaxin Fan, Xiang Wan, Feng Jiang, and Benyou Wang. 2024.
\newblock {PlatoLM}: Teaching {LLMs} in multi-round dialogue via a user simulator.
\newblock In \emph{Proceedings of the 62nd Annual Meeting of the Association for Computational Linguistics (ACL'24)}.

\bibitem[{Lee et~al.(2023)Lee, Srivastava, Hardy, Thickstun, Durmus, Paranjape, Gerard-Ursin, Li, Ladhak, Rong, Wang, Kwon, Park, Cao, Lee, Bommasani, Bernstein, and Liang}]{lee2023interaction}
Mina Lee, Megha Srivastava, Amelia Hardy, John Thickstun, Esin Durmus, Ashwin Paranjape, Ines Gerard-Ursin, Xiang~Lisa Li, Faisal Ladhak, Frieda Rong, Rose~E. Wang, Minae Kwon, Joon~Sung Park, Hancheng Cao, Tony Lee, Rishi Bommasani, Michael Bernstein, and Percy Liang. 2023.
\newblock Evaluating human-language model interaction.
\newblock \emph{Transactions on Machine Learning Research}.

\bibitem[{Li et~al.(2024{\natexlab{a}})Li, Li, Wang, and Du}]{li2024iqaeval}
Ruosen Li, Ruochen Li, Barry Wang, and Xinya Du. 2024{\natexlab{a}}.
\newblock {IQA-EVAL}: Automatic evaluation of human-model interactive question answering.
\newblock In \emph{Proceedings of the 38th International Conference on Neural Information Processing Systems (NeurIPS'24)}.

\bibitem[{Li et~al.(2024{\natexlab{b}})Li, Balachandran, Feng, Ilgen, Pierson, Koh, and Tsvetkov}]{li2024mediq}
Shuyue~Stella Li, Vidhisha Balachandran, Shangbin Feng, Jonathan~S. Ilgen, Emma Pierson, Pang~Wei Koh, and Yulia Tsvetkov. 2024{\natexlab{b}}.
\newblock Mediq: Question-asking {LLMs} and a benchmark for reliable interactive clinical reasoning.
\newblock In \emph{Proceedings of the 38th International Conference on Neural Information Processing Systems (NeurIPS'24)}.

\bibitem[{Li et~al.(2024{\natexlab{c}})Li, Chiang, Frick, Dunlap, Zhu, Gonzalez, and Stoica}]{li2024arenahard}
Tianle Li, Wei-Lin Chiang, Evan Frick, Lisa Dunlap, Banghua Zhu, Joseph~E. Gonzalez, and Ion Stoica. 2024{\natexlab{c}}.
\newblock From live data to high-quality benchmarks: The arena-hard pipeline.
\newblock \url{https://lmsys.org/blog/2024-04-19-arena-hard/}.

\bibitem[{Liang et~al.(2023)Liang, Bommasani, Lee, Tsipras, Soylu, Yasunaga, Zhang, Narayanan, Wu, Kumar et~al.}]{liang2023helm}
Percy Liang, Rishi Bommasani, Tony Lee, Dimitris Tsipras, Dilara Soylu, Michihiro Yasunaga, Yian Zhang, Deepak Narayanan, Yuhuai Wu, Ananya Kumar, et~al. 2023.
\newblock Holistic evaluation of language models.
\newblock \emph{Transactions of Machine Learning Research}.

\bibitem[{Lin et~al.(2024)Lin, Deng, Chandu, Brahman, Ravichander, Pyatkin, Dziri, Bras, and Choi}]{lin2024wildbench}
Bill~Yuchen Lin, Yuntian Deng, Khyathi Chandu, Faeze Brahman, Abhilasha Ravichander, Valentina Pyatkin, Nouha Dziri, Ronan~Le Bras, and Yejin Choi. 2024.
\newblock {WildBench}: Benchmarking {LLMs} with challenging tasks from real users in the wild.
\newblock In \emph{arXiv preprint arXiv:2406.04770}.

\bibitem[{{Llama Team, AI@Meta}(2024)}]{meta2024llama3}
{Llama Team, AI@Meta}. 2024.
\newblock The llama 3 herd of models.
\newblock \emph{arXiv preprint arXiv:2407.21783}.

\bibitem[{Miller(2024)}]{miller2024adding}
Evan Miller. 2024.
\newblock Adding error bars to evals: A statistical approach to language model evaluations.
\newblock \emph{arXiv preprint arXiv:2411.00640}.

\bibitem[{OpenAI(2023)}]{openai2023gpt4}
OpenAI. 2023.
\newblock {GPT-4} technical report.
\newblock \emph{arXiv preprint arXiv:2303.08774}.

\bibitem[{Park et~al.(2023)Park, O'Brien, Cai, Morris, Liang, and Bernstein}]{park2023generative}
Joon~Sung Park, Joseph~C. O'Brien, Carrie~J. Cai, Meredith~Ringel Morris, Percy Liang, and Michael~S. Bernstein. 2023.
\newblock Generative agents: Interactive simulacra of human behavior.
\newblock In \emph{Proceedings of the 36th Annual ACM Symposium on User Interface Software and Technology (UIST'23)}.

\bibitem[{Posch et~al.(2022)Posch, Bleier, Flöck, Lechner, Kinder-Kurlanda, Helic, and Strohmaier}]{posch2022crowd}
Lisa Posch, Arnim Bleier, Fabian Flöck, Clemens~M. Lechner, Katharina Kinder-Kurlanda, Denis Helic, and Markus Strohmaier. 2022.
\newblock Characterizing the global crowd workforce:a cross-country comparison of crowdworker demographics.
\newblock \emph{Human Computation}, 9(1):22--57.

\bibitem[{Ren et~al.(2024)Ren, Qiu, Qu, Liu, Zhao, Wu, Wen, and Wang}]{ren2024bases}
Ruiyang Ren, Peng Qiu, Yingqi Qu, Jing Liu, Xin Zhao, Hua Wu, Ji-Rong Wen, and Haifeng Wang. 2024.
\newblock {BASES}: Large-scale web search user simulation with large language model based agents.
\newblock In \emph{Findings of the Association for Computational Linguistics: EMNLP 2024}.

\bibitem[{Shao et~al.(2024)Shao, Samuel, Jiang, Yang, and Yang}]{shao2024cogym}
Yijia Shao, Vinay Samuel, Yucheng Jiang, John Yang, and Diyi Yang. 2024.
\newblock Collaborative gym: A framework for enabling and evaluating human-agent collaboration.
\newblock \emph{arXiv preprint arXiv:2412.15701}.

\bibitem[{Suh et~al.(2025)Suh, Jahanparast, Moon, Kang, and Chang}]{suh2025subpop}
Joseph Suh, Erfan Jahanparast, Suhong Moon, Minwoo Kang, and Serina Chang. 2025.
\newblock Language model fine-tuning on scaled survey data for predicting distributions of public opinions.
\newblock \emph{arXiv preprint arXiv:2502.16761}.

\bibitem[{{U.S. Bureau of Labor Statistics}(2025)}]{bureau2025labor}
{U.S. Bureau of Labor Statistics}. 2025.
\newblock {Labor Force Statistics from the Current Population Survey}.
\newblock \url{https://www.bls.gov/web/empsit/cpseea06.htm}. Accessed March 28, 2025.

\bibitem[{{U.S. Census Bureau}(2024)}]{census2024quickfacts}
{U.S. Census Bureau}. 2024.
\newblock {National Population by Characteristics: 2020-2024}.
\newblock \url{https://www.census.gov/data/tables/time-series/demo/popest/2020s-national-detail.html}. Accessed March 28, 2025.

\bibitem[{Vaccaro et~al.(2024)Vaccaro, Almaatouq, and Malone}]{vaccaro2024combo}
Michelle Vaccaro, Abdullah Almaatouq, and Thomas Malone. 2024.
\newblock When combinations of humans and {AI} are useful: A systematic review and meta-analysis.
\newblock \emph{Nature Human Behaviour}, 8:2293--2303.

\bibitem[{Wang et~al.(2025)Wang, Morgenstern, and Dickerson}]{wang2024flatten}
Angelina Wang, Jamie Morgenstern, and John~P. Dickerson. 2025.
\newblock Large language models that replace human participants can harmfully misportray and flatten identity groups.
\newblock \emph{Nature Machine Intelligence}.

\bibitem[{Yang et~al.(2018)Yang, Qi, Zhang, Bengio, Cohen, Salakhutdinov, and Manning}]{yang2018hotpotqa}
Zhilin Yang, Peng Qi, Saizheng Zhang, Yoshua Bengio, William~W. Cohen, Ruslan Salakhutdinov, and Christopher~D. Manning. 2018.
\newblock {HotpotQA}: A dataset for diverse, explainable multi-hop question answering.
\newblock In \emph{Proceedings of the 2018 Conference on Empirical Methods in Natural Language Processing (EMNLP'18)}.

\bibitem[{Zhao et~al.(2024)Zhao, Ren, Hessel, Cardie, Choi, and Deng}]{zhao2024wildchat}
Wenting Zhao, Xiang Ren, Jack Hessel, Claire Cardie, Yejin Choi, and Yuntian Deng. 2024.
\newblock {WildChat}: 1m chatgpt interaction logs in the wild.
\newblock In \emph{Proceedings of the 12th International Conference on Learning Representations (ICLR'24)}.

\bibitem[{Zheng et~al.(2024)Zheng, Chiang, Sheng, Li, Zhuang, Wu, Zhuang, Li, Lin, Xing, Gonzalez, Stoica, and Zhang}]{zheng2024lmsys}
Lianmin Zheng, Wei-Lin Chiang, Ying Sheng, Tianle Li, Siyuan Zhuang, Zhanghao Wu, Yonghao Zhuang, Zhuohan Li, Zi~Lin, Eric~P. Xing, Joseph~E. Gonzalez, Ion Stoica, and Hao Zhang. 2024.
\newblock {LMSYS-Chat-1M}: A large-scale real-world {LLM} conversation dataset.
\newblock In \emph{Proceedings of the 12th International Conference on Learning Representations (ICLR'24)}.

\bibitem[{Zheng et~al.(2023)Zheng, Chiang, Sheng, Zhuang, Wu, Zhuang, Lin, Li, Li, Xing, Zhang, Gonzalez, and Stoica}]{zheng2023llmjudge}
Lianmin Zheng, Wei-Lin Chiang, Ying Sheng, Siyuan Zhuang, Zhanghao Wu, Yonghao Zhuang, Zi~Lin, Zhuohan Li, Dacheng Li, Eric Xing, Hao Zhang, Joseph~E. Gonzalez, and Ion Stoica. 2023.
\newblock Judging {LLM}-as-a-judge with mt-bench and chatbot arena.
\newblock In \emph{Proceedings of the 37th International Conference on Neural Information Processing Systems (NeurIPS'23)}.

\end{thebibliography}
